\documentclass[9pt,conference]{IEEEtran}
\IEEEoverridecommandlockouts % show footnote comments
\usepackage{blkarray}                                      % to support matrix
\usepackage{algpseudocode}                                 % new algorithm package
\usepackage{algorithm}
\usepackage{graphicx}                                      % include pdf figures
\usepackage{amsmath}
\usepackage{amssymb}
\usepackage{amsfonts}
\usepackage{amsthm}
\usepackage[mathcal]{eucal}
\usepackage{mathrsfs}
\usepackage{booktabs}
\usepackage{enumerate}
\usepackage{multirow}
\usepackage[subrefformat=parens,farskip=0pt,justification=centering]{subfig}
\usepackage{color}
\usepackage{cite}                                          % more citations in one bracket
\usepackage{comment}                                       % use comment
\usepackage{soul}                                          % use highlight command \hl{}
\soulregister\cite7
\soulregister\ref7
\soulregister\pageref7
\usepackage{etoolbox}                                      % commands \newtoggle, \toggletrue, \iftoggle
\usepackage{url}
\usepackage{nth}                                           % nth command
\usepackage{bm}                                            % bm command
\usepackage{courier}
\usepackage{balance}
\usepackage{threeparttable}
\usepackage{xcolor,colortbl}
\usepackage{footnote}
\usepackage[bookmarks=false]{hyperref}
\hypersetup{
    colorlinks = true,
    citecolor  = blue,
    linkcolor  = blue,
    urlcolor   = blue,
}
\usepackage{tikz}
\usetikzlibrary{patterns,snakes}
\usetikzlibrary{positioning,calc,fit,decorations.pathmorphing,shapes.geometric, shapes.gates.logic.US, calc}
\usetikzlibrary{arrows,arrows.meta,decorations.markings,shapes,shapes.arrows}
\usetikzlibrary{decorations,decorations.pathreplacing}
\usetikzlibrary{backgrounds}
\usepackage{filecontents}                                  % support to pgfplots
\usepackage{pgfplots}
\usepackage{pgfplotstable}
\usepackage{scalefnt}
\pgfplotsset{compat=newest}
\usepackage{caption}
\usepackage{cleveref}
\Crefformat{figure}{Fig.~#2#1#3}                           % "Fig.", instead of "Figure"
\Crefname{subfigure}{Fig.}{Figs.}
\Crefname{figure}{Fig.}{Figs.}
\Crefformat{table}{TABLE~#2#1#3}                           % "TABLE", instead of "Table"
\usepackage[figuresright]{rotating}
\definecolor{CUHKorange}{RGB}{244,106,18} %F47012
\definecolor{CUHKblue}{RGB}{0,111,190}    %006FBE
\definecolor{CUHKgreen}{RGB}{0,127,128}   %007F80
\definecolor{CUHKred}{RGB}{228,46,36}     %E42E24
\definecolor{CUHKyellow}{RGB}{198,148,34} %C69422
\definecolor{CUHKdark}{RGB}{114,44,114}   %722C72
\definecolor{CUHKmiddle}{RGB}{144,44,144} %902C90
\definecolor{CUHKlight}{RGB}{167,44,167} 
\definecolor{CUHKpurple}{RGB}{117,15,109}
\definecolor{CUHKgold}{RGB}{221,163,0}
\definecolor{CUHKribbon}{RGB}{244,223,176}
\definecolor{CUHKblack}{RGB}{34,24,21}

\renewcommand{\vec}[1]{\boldsymbol{#1}}    % re-define vec command

\usepackage{tcolorbox}
\tcbuselibrary{skins,breakable}
    {\endtcolorbox}

\usepackage[top=0.75in,bottom=0.80in,left=0.58in,right=0.58in]{geometry}
\crefname{mytheorem}{Theorem}{Theorems}
\crefname{mylemma}{Lemma}{Lemmas}
\crefname{myclaim}{Claim}{Claims}
\crefname{myproperty}{Property}{Properties}
\crefname{mycorollary}{Corollary}{Corollaries}

\algrenewcommand\textproc{\texttt}

\makeatletter
\let\OldStatex\Statex
\renewcommand{\Statex}[1][3]{%
  \setlength\@tempdima{\algorithmicindent}%
  \OldStatex\hskip\dimexpr#1\@tempdima\relax
}
\makeatother

\RequirePackage[normalem]{ulem} %DIF PREAMBLE
\RequirePackage{color}\definecolor{RED}{rgb}{1,0,0}\definecolor{BLUE}{rgb}{0,0,1} %DIF PREAMBLE
\usepackage{times}
\usepackage{algpseudocodex}
\usepackage{upgreek}
\usepackage[export]{adjustbox}

\graphicspath{{./figs/}}

\newcommand{\etal}{\textit{et al}. }
\newcommand{\ie}{\textit{i}.\textit{e}., }

\newcommand{\fname}{Nitho}
\begin{document}
\date{}

\title{
Physics-Informed Optical Kernel Regression Using Complex-valued Neural Fields
}

\iftrue
\author{
    Guojin Chen$^1$, \quad
    Zehua Pei$^1$, \quad
    Haoyu Yang$^2$, \quad
    Yuzhe Ma$^3$, \quad
    Bei Yu$^1$, \quad
    Martin Wong$^1$\\
    $^1$Chinese University of Hong Kong \quad
    $^2$nVIDIA \quad
    $^3$HKUST(GZ)
    %{\tt\small \{gjchen21,byu\}@cse.cuhk.edu.hk}
}
\fi

\iffalse
\author{Guojin Chen}{https://www.overleaf.com/project/6353935b965fa750e35f0657}
\affiliation{
    \institution{Chinese University of Hong Kong}
}
\email{gjchen21@cse.cuhk.edu.hk}

\author{Bei Yu}
\affiliation{
    \institution{Chinese University of Hong Kong}
}
\email{byu@cse.cuhk.edu.hk}
\fi

\maketitle

\begin{abstract}
Lithography is fundamental to integrated circuit fabrication, necessitating large computation overhead.
The advancement of machine learning (ML)-based lithography models alleviates the trade-offs between manufacturing process expense and capability.
However, all previous methods regard the lithography system as an image-to-image black box mapping,
utilizing network parameters to learn by rote mappings from massive mask-to-aerial or mask-to-resist image pairs, resulting in poor generalization capability.
In this paper, we propose a new ML-based paradigm disassembling the rigorous lithographic model into non-parametric mask operations
and learned optical kernels containing determinant source, pupil, and lithography information.
By optimizing complex-valued neural fields to perform optical kernel regression from coordinates,
our method can accurately restore lithography system using a small-scale training dataset with fewer parameters,
demonstrating superior generalization capability as well.
Experiments show that our framework can use 31\% of parameters while achieving 69$\times$ smaller mean squared error
with 1.3$\times$ higher throughput than the state-of-the-art.
\end{abstract}
\section{Introduction}
\label{sec:intro}

\begin{figure}
  \centering
  \subfloat[]{ \includegraphics[width=.56\linewidth]{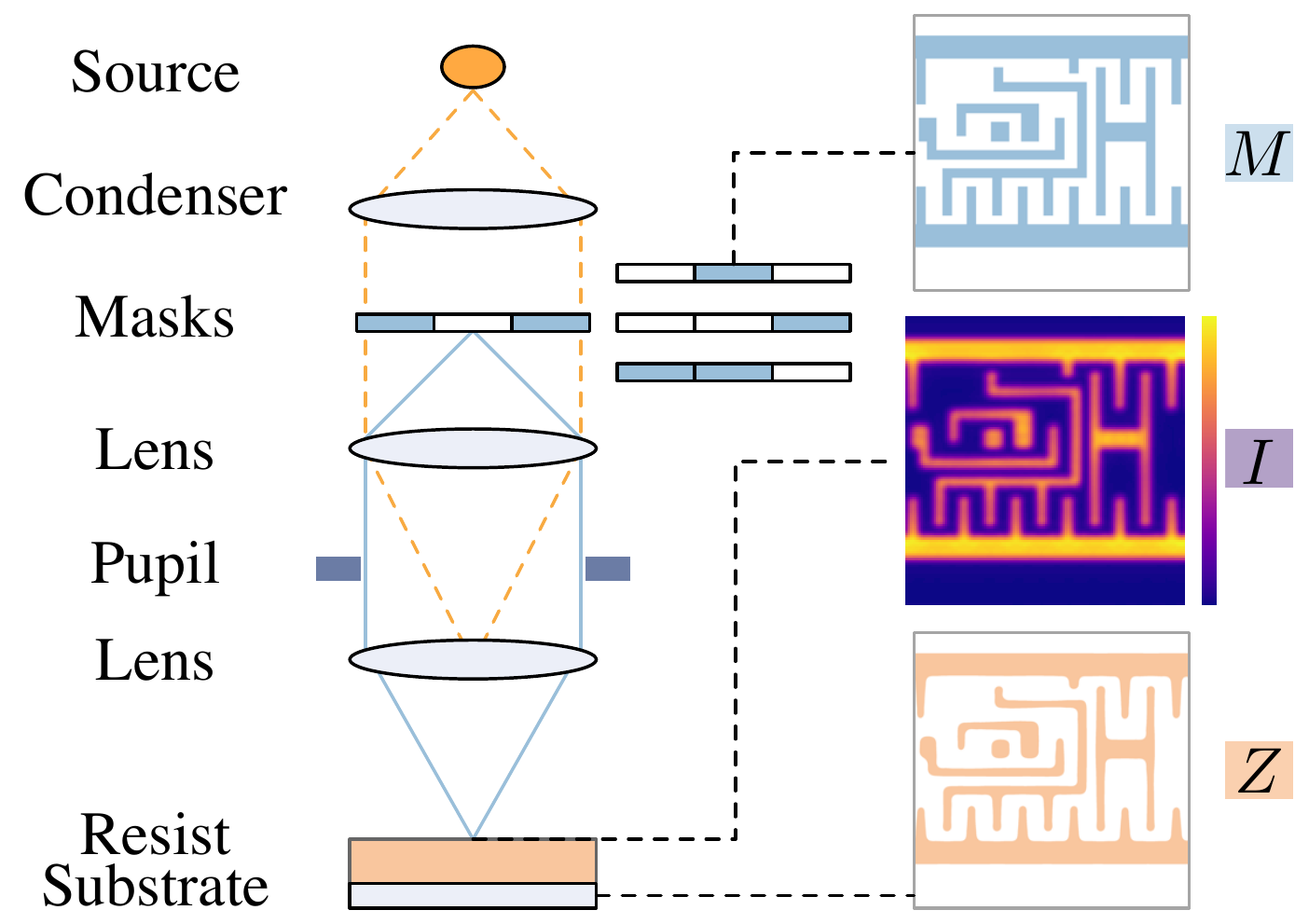} \label{fig:lens}}
  \subfloat[]{ \includegraphics[width=.28\linewidth]{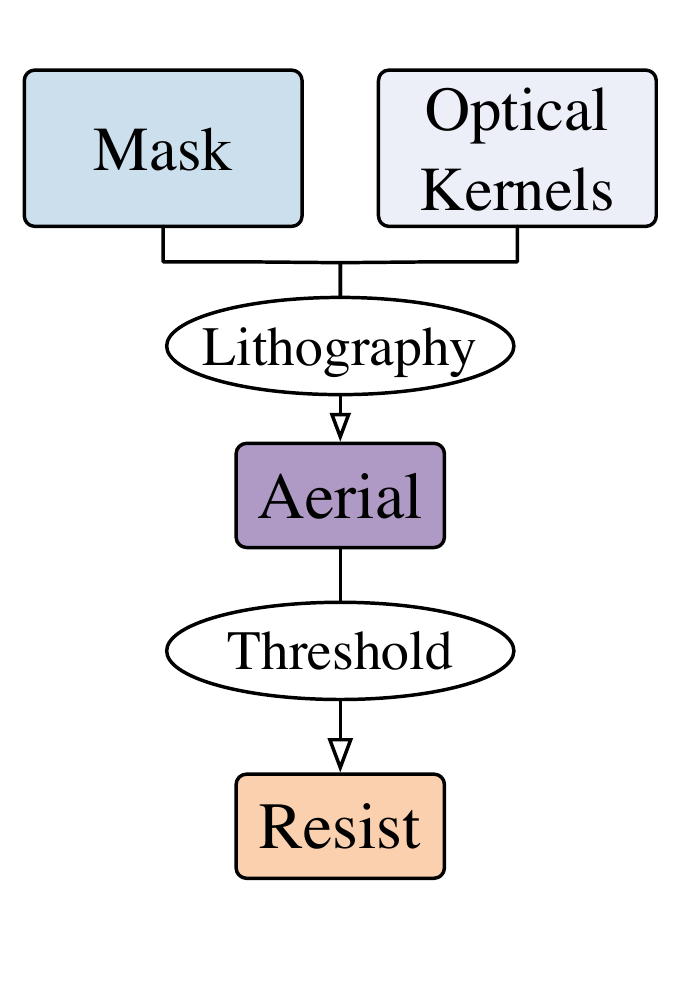} \label{fig:litho_flow}}
  \caption{
    (a) Components of the lithography imaging system: illumination source, lenses, and pupil.
    (b) Lithography simulation flow using source- and pupil-dependent optical kernels.
  }
  \label{fig:litho}
\end{figure}

In modern chip manufacturing, lithography simulation is one of the most critical technologies which affects many other fabrication processes. %And the importance grows as the feature size continuously shrinks.
Traditional rigorous lithography methods model optical lithography numerically by multifold integrals of the mutual intensity of all contributing source points through the projection system from the illuminator~\cite{DFM-B2008-Mack}, which is computationally expensive and challenging, even when equipped with approximation algorithms.
Given a mask, lithography simulation generates an aerial image that is defined as the intensity distribution at the wafer plane, as depicted in \Cref{fig:litho}.
An estimate of the binary resist image can be obtained by applying an exposure-dose-dependent intensity threshold on an aerial image.

Considering the significance and challenges of lithography simulation, enormous advances in computing power and machine learning (ML) algorithms stimulate the growth of exploiting data-driven methods for lithography modeling.
Generally, ML-based methods apply groups of transformations and learnable parameters to fit a mask-to-aerial~\cite{ISPD-2020-TEMPO}  or mask-to-resist~\cite{DFM-DAC2019-Ye,OPC-TCAD-DAMO,DAC22-DOINN-Yang} mapping.
LithoGAN~\cite{DFM-DAC2019-Ye} is a lithography framework using conditional generative adversarial nets (cGAN) to predict the resist pattern of a cropped clip.
LithoGAN introduces an extra branch of convolution neural networks (CNN) to perform center coordinates regression for better accuracy, limiting it is only applicable to a single contact.
DAMO~\cite{OPC-TCAD-DAMO} exploits the UNet++ as the deep convolutional generative adversarial networks generator to engage the deep lithography simulator in a high-resolution manner, which can be further extended to processing full-chip scale.
Yang~\etal proposes a dual-band optics-inspired neural network (DOINN)~\cite{DAC22-DOINN-Yang} lithography model with Fourier Neural Operator (FNO) to properly leverage both high-frequency and global low-frequency components of mask-resist pairs.
The state-of-the-art (SOTA) aerial image model, TEMPO~\cite{ISPD-2020-TEMPO}, applies cGAN as a thick mask effect modeling framework capable of predicting 3D aerial images at different heights.

\begin{figure}
  \centering
  \includegraphics[width=.9\linewidth]{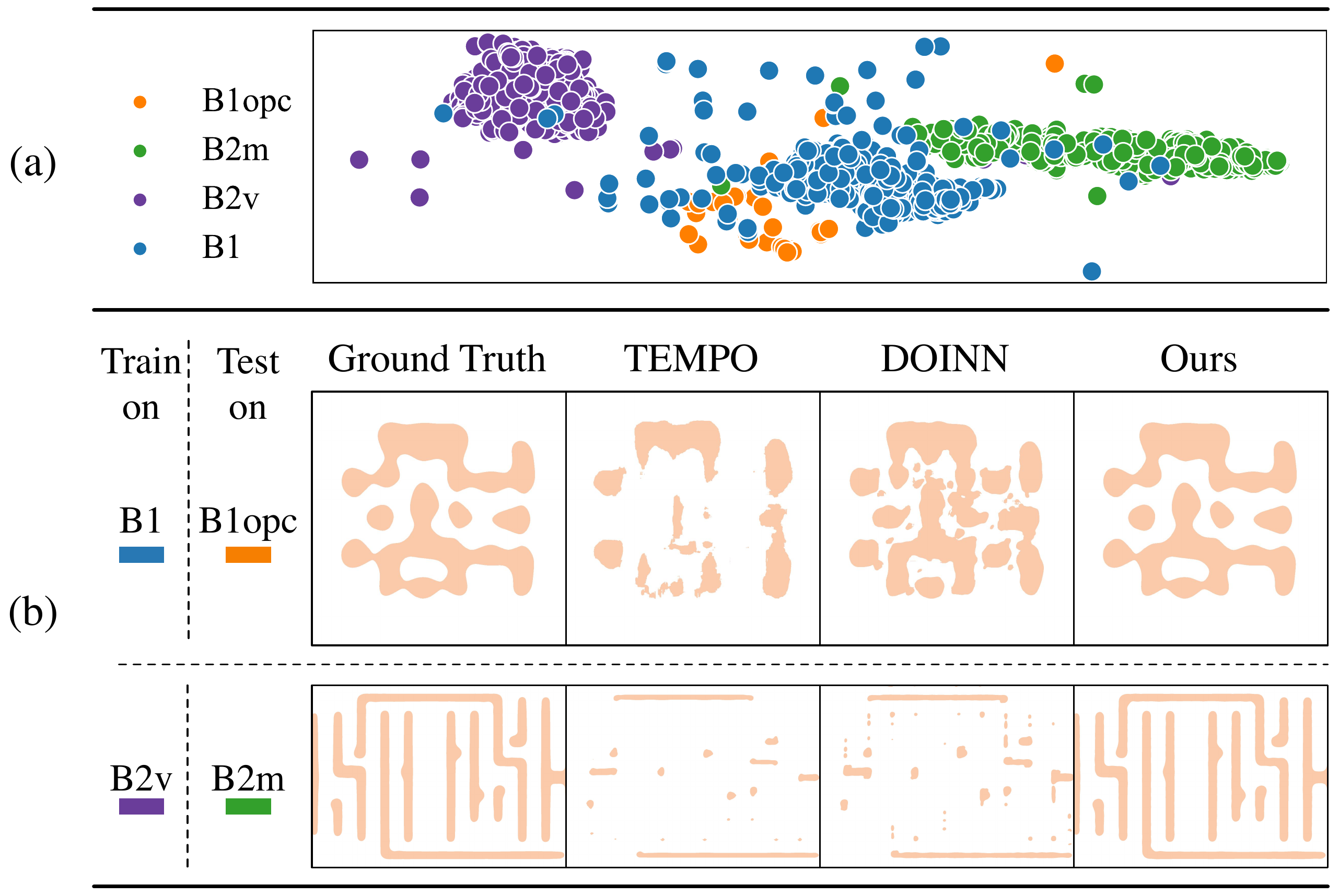}
  \caption{
    (a) t-SNE distribution of datasets listed in \Cref{tab:data}.
    (b) Comparison of generalization capability on out-of-distribution datasets.}
  \label{fig:general}
\end{figure}

However, all these CNN-based image generation models still have some drawbacks for lithography modeling problems.
Since all previous arts are fitting a particular image-to-image mapping from massive training pairs instead of learning the true lithography behavior,
the learned parameters have a strong bias on the mask shapes, layer types, and training dataset distributions,
which fails to generalize on out-of-distribution (OOD) datasets.
In \Cref{fig:general}(a), we visualize the t-Distributed Stochastic Neighbor Embedding (t-SNE) of four datasets used in this work and DOINN~\cite{DAC22-DOINN-Yang}.
As depicted in \Cref{fig:general}(b), despite being well-trained, DOINN and TEMPO fail to generalize on OOD datasets.
A possible solution is to use larger models to fit different distributions.
However, it will increase computational cost exponentially while suffering from precision loss, thus is not a one-fit-all approach.

The generalization failure of a DNN model for lithography modeling will not be desired, given the fact that a real simulator can handle masks of different layers without any degradation.
If we revisit the lithography simulation model, as depicted in \Cref{fig:litho_flow}, it can be noticed that the intrinsic knowledge of a lithography system is in the optical kernel, which is almost ignored when conducting image-to-image mapping based on a neural network.
It motivates us to investigate the mystery of the optical kernel for fast and accurate lithography modeling.
The rigorous Hopkins lithography model~\cite{hopkins1953diffraction} separates the influence of the mask and the lithographic imaging system, including pupil function and illumination.
The latter is an intermediate variable often referred to as transmission cross-coefficient (TCC) $\mathcal{T}$,
which can be pre-computed and stored as optical kernels to improve the computing efficiency~\cite{Fhner2014ArtificialEF}.
Now we may ask: \textit{Can we train a deep neural network to replace the optical kernels for efficient lithography modeling?}
Nevertheless, this framework encounters several obstacles.
First, TCC kernels are hard to obtain and calibrate, making direct regression impractical.
We are seeking for a feasible alternative to decode the intermediate optical kernels from final imaging samples of mask images and aerial images.
Second, TCC-related computations are performed in the spatial frequency domain with complex values, which implies that the network needs to support complex-valued computations.

To handle the above issues, we attempt to use physical \textit{`resolution limits'} to design the TCC optical kernel dimensions.
Then, a set of differentiable complex-valued neuron layers are designed and implemented.
Last, optical kernels are location-dependent matrices.
Inspired by \textit{neural radiance field} (NeRF)~\cite{ECCV-2020-NeRF},
we design and implement a complex-valued multilayer perception ($\operatorname{\mathbb{C}MLP}$),
which only takes coordinates as input to perform optical kernel regression.
Moreover, a novel training procedure that separates mask processing and optical kernel prediction is proposed. % to generate the final aerial image  in \Cref{subsec:forward_training}.
To conclude, in this paper, we propose Nitho, a NeRF-inspired and physics-informed lithographic network optimizing complex-valued neural fields to predict the TCC spectrum as optical kernels,
which will be further multiplied with the mask spectrum to generate a high-accuracy aerial image.

The major contributions of this paper include:
\begin{enumerate}
  \item To the best of our knowledge, it is the first neural network-based framework predicting TCCs that models the true lithography behavior instead of learning an image-to-image black box mapping.
  \item Inspired by Hopkins model, Nitho abandons CNN-based architecture and creates a new training paradigm separating the influence of mask and lithography system using a simple coordinate-based complex-valued multilayer perception.
  \item Similar to industrial simulator, the physics-informed design maximizes Nitho's generalization capability on mixed types of mask data with no more specific treatment on different types of masks.
  \item The proposed Nitho framework can use 31\% model size on less training data to generate high-resolution aerial images while achieving $69\times$ smaller mean squared error and higher accuracy than SOTA.
Compared with traditional lithography simulator, Nitho can achieve $90\times$ speedup with less than 1\% accuracy loss.
\end{enumerate}

\section{Preliminaries}
\label{sec:prelim}
This section is about basic terminology related to \fname~framework and problem formulation.
Throughout this paper, we use lowercase letters for scalars, bold lowercase for vectors, and bold uppercase for matrices,
$\odot$ for element-wise production, $^{*}$ for complex conjugate,
$\vec{I}$, $\vec{Z}$, $\vec{M}$ for aerial, resist and mask image,
$\mathcal{F}$ and $\mathcal{F}^{-1}$ for fast Fourier transform (FFT) and inverse FFT, respectively

\subsection{Hopkins Model and Transmission Cross-Coefficient (TCC)}
In 1953, Hopkins~\cite{hopkins1953diffraction} developed a formulation that has been wildly used afterward,
whose purpose is a separation of the influence of the mask and the imaging system, including the pupil function and the illumination.
For numerical convenience, Hopkins method is often stated in terms of the spatial spectrum of the aerial intensity $\vec{I}$:
% in the image plane:
\begin{equation}
  \begin{aligned}
  \mathcal{F}(\vec{I})(f, g)=\iint_{-\infty}^{\infty} \mathcal{T}\left(\left(f^{\prime}+f, g^{\prime}+g\right),\left(f^{\prime}, g^{\prime}\right)\right) \\
  \mathcal{F}(\vec{M})\left(f^{\prime}+f, g^{\prime}+g\right) \mathcal{F}(\vec{M})^*\left(f^{\prime}, g^{\prime}\right) \mathrm{d} f^{\prime} \mathrm{d} g^{\prime},
  \end{aligned}
  \label{eq:hopkins}
\end{equation}
where $\vec{M}$ is mask, $(f, g)$ is its frequencies. $\mathcal{T}$ is TCC given by:
\begin{equation}
  \label{eq:tcc}
  % \small{
    \begin{aligned}
      \mathcal{T}\left(\left(f^{\prime}, g^{\prime}\right),\left(f^{\prime \prime}, g^{\prime \prime}\right)\right):=\iint_{-\infty}^{\infty} \mathcal{F}(J)(f, g) \\
      \mathcal{F}(H)\left(f+f^{\prime}, g+g^{\prime}\right) \mathcal{F}(H)^*\left(f+f^{\prime \prime}, g+g^{\prime \prime}\right) \mathrm{d} f \mathrm{~d} g,
    \end{aligned}
  % }
\end{equation}
where the weight factor $J$ solely depends on effective source, $H$ is projector transfer function.
Observing \Cref{eq:hopkins,eq:tcc}, $\mathcal{T}$ is independent of mask transmission functions $\mathcal{F}(\vec{M})$.
For applications where source and pupil are fixed, but images must be calculated for a wide range of different masks,
the TCC matrix can be pre-calculated and stored to accelerate imaging calculations.

\subsubsection{Sum of Coherent Sources Approach (SOCS)}
\label{subsubsec:socs}
The TCC is Hermitian and positive definite and can be decomposed into a set of eigenvalues with corresponding eigenvectors.
An approximation solution for the Hopkins imaging equations called \textit{Sum of Coherent Source} (SOCS)~\cite{cobb1995sum} using SVD algorithm which provides
relatively fast computation times and is thus widely used in some inverse imaging calculation tasks such as mask optimization.
The TCC spectrum matrix can be written as
\begin{equation}
  \mathcal{F}(\mathcal{T})=\sum_i \alpha_i \mathbf{h}_i \mathbf{h}_i^* ,
  \label{eq:socs_svd}
\end{equation}
where $\mathbf{h}_i$ are the eigenvectors and $\alpha_i$ are eigenvalues of TCC spectrum $\mathcal{F}(\mathcal{T})$.
By Fourier-back-transfoming we then obtain the SOCS formula:
\begin{equation}
  \label{eq:socs}
  \vec{I}=\sum_{i} \alpha_i\left|\mathcal{F}^{-1}\left(\mathcal{F}\left(\boldsymbol{h}_i\right) \odot \mathcal{F}(\boldsymbol{M})\right)\right|^2 .
\end{equation}

\subsection{Problem Formulation}
In aerial image generation stage, the lithography modeling can be viewed as a pixel-wise regression task,
therefore we adopt Mean of Squared Error (MSE), Peak Signal-to-Noise Ratio (PSNR), Max Error (ME) to evaluate the performance.
In resist image generation stage, it can be viewed as a pixel-wise classification problem,
we adopt the same matrices including Mean Intersection Over Union (mIOU) and Mean Pixel Accuracy (mPA) as used in SOTA~\cite{DAC22-DOINN-Yang}.

\subsubsection{MSE, PSNR.}
Given aerial image $\vec{I}$ and its prediction $\hat{\vec{I}}$, the MSE can be calculated as
\begin{equation}
  \operatorname{MSE}(\vec{I}, \hat{\vec{I}}) = \frac{1}{N}\sum_{i}^{N}(\vec{I}_i - \hat{\vec{I}}_i)^2,
\end{equation}
where $N$ is the total number of pixels in $\vec{I}$ and $\hat{\vec{I}}$. The smaller MSE, the better performance.
PSNR is a quality measurement between the original and the predicted image, in decibels.
The higher PSNR, the better quality of the predicted image. PSNR is given as
\begin{equation}
  \operatorname{PSNR}(\vec{I}, \hat{\vec{I}}) = 10 * \log_{10} ({\operatorname{max}(\vec{I})^2} / {\operatorname{MSE}(\vec{I}, \hat{\vec{I}})}).
\end{equation}

\subsubsection{mIOU, mPA}
Given k classes of predicted resist patterns $\hat{\vec{Z}}_i$ and their ground truth $\vec{Z}_i$, $i = 1, 2 \ldots k$, the mIOU and mPA are:
\begin{equation}
  \small
  \begin{aligned}
    \operatorname{mIOU}(\vec{Z}, \hat{\vec{Z}})=\frac{1}{k} \sum_{i=1}^k \frac{\vec{Z}_i \cap \hat{\vec{Z}}_i}{\vec{Z}_i \cup \hat{\vec{Z}}_i},~
    \operatorname{mPA}(\vec{Z}, \hat{\vec{Z}})=\frac{1}{k} \sum_{i=1}^k \frac{\hat{\vec{Z}}_i \cap \vec{Z}_i}{\vec{Z}_i} .
  \end{aligned}
\end{equation}

\subsubsection{ME}
Max error (ME) is  measured by the maximum difference between the predicted aerial image $\hat{\vec{I}}$ and its ground truth $\vec{I}$:
\begin{equation}
  \operatorname{ME} = \operatorname{Max}(|\vec{I} - \hat{\vec{I}}|) .
\end{equation}

\subsubsection{Problem definition}
Given a set of mask images $\mathcal{M}_{tr} = \{\vec{M}_1, \ldots, \vec{M}_n\}$
and their corresponding aerial images $\mathcal{I}_{tr} = \{\vec{I}_1, \ldots, \vec{I}_n\}$.
Resist images $\mathcal{Z}_{tr} = \{\vec{Z}_1, \ldots, \vec{Z}_n\}$ can be obtained by a constant threshold $I_{thres}$.
Our target is to design a machine learning model that can
reconstruct the TCC optical kernels, then for new designs $\mathcal{M}_{te}$, $\mathcal{I}_{te}$, $\mathcal{Z}_{te}$
using the SOCS formula in \Cref{eq:socs} to get the predicted aerial images $\hat{\mathcal{I}} = \{\hat{\vec{I}}_1, \ldots, \hat{\vec{I}}_n\}$,
and predicted resist images $\hat{\mathcal{Z}}$.
The $\operatorname{MSE}(\mathcal{I}_{te}, \hat{\mathcal{I}})$, $\operatorname{ME}(\mathcal{I}_{te}, \hat{\mathcal{I}})$ can be minimized
and $\operatorname{PSNR}(\mathcal{I}_{te}, \hat{\mathcal{I}})$,
$\operatorname{mIOU}(\mathcal{Z}_{te}, \hat{\mathcal{Z}})$, $\operatorname{mPA}(\mathcal{Z}_{te}, \hat{\mathcal{Z}})$ can be maximized.

\begin{figure*}
  % \vspace{-1em}
  \centering
  \subfloat[]{ \includegraphics[width=.76\linewidth]{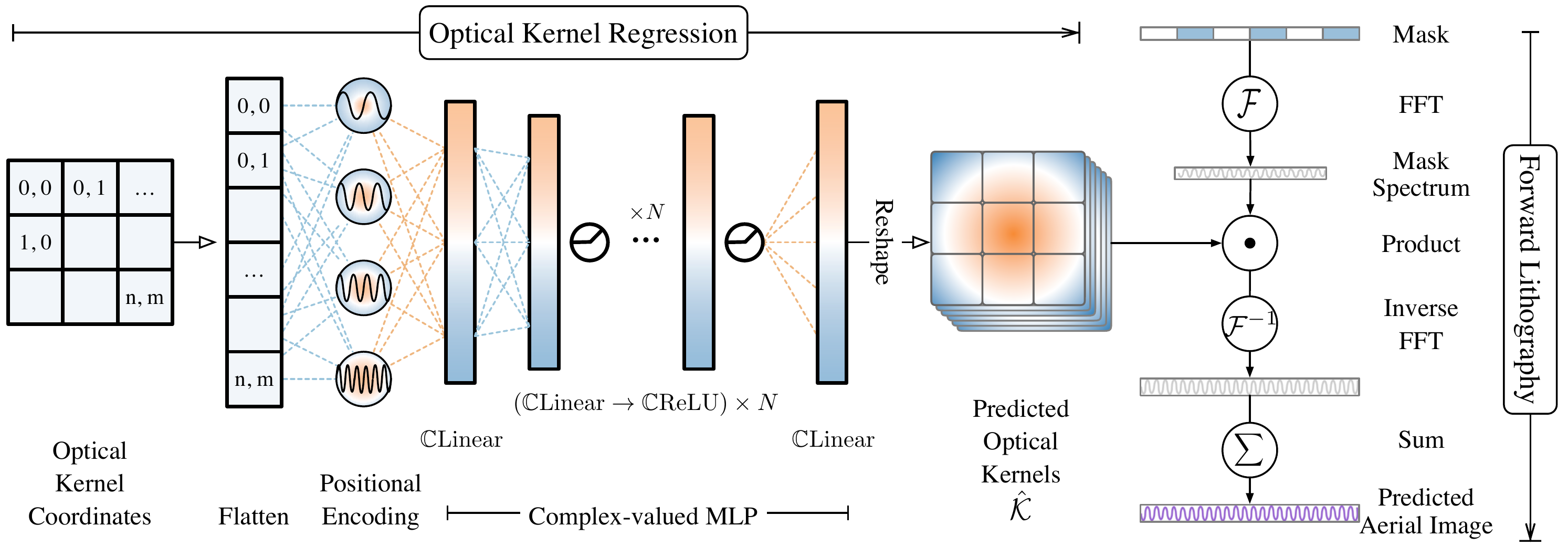} \label{fig:nitho}}
  \subfloat[]{ \includegraphics[width=.19\linewidth]{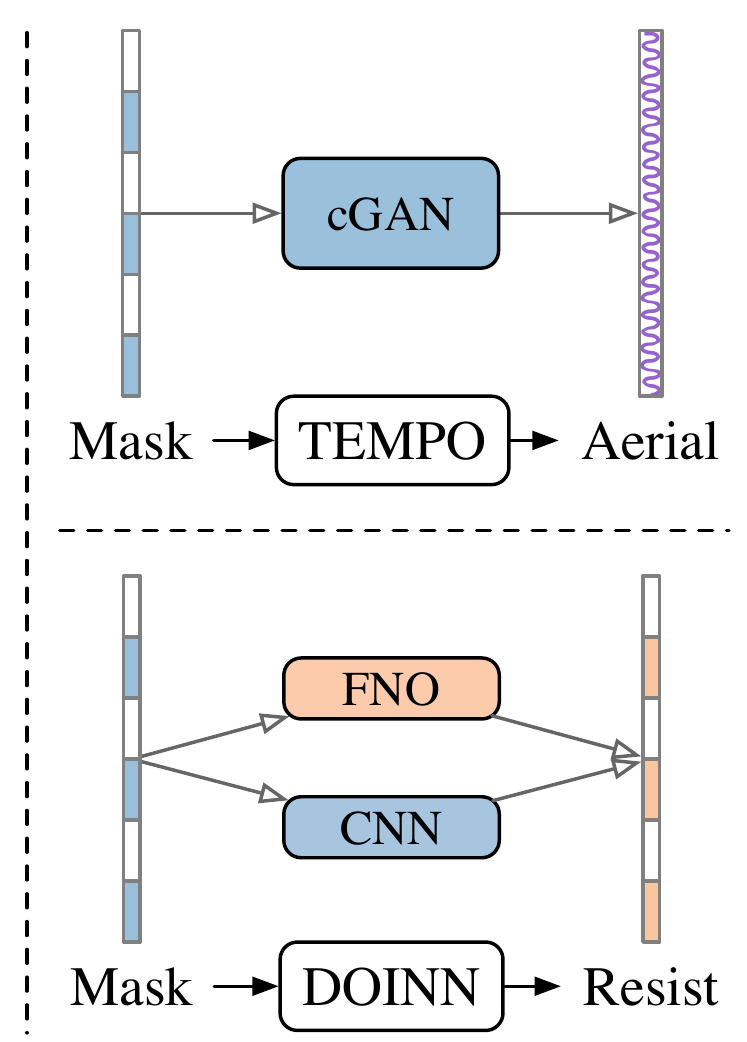} \label{fig:prev}}
  \caption{
    (a) The overall aerial image prediction pipeline of Nitho framework, which separates mask-related linear operations from optical kernel regression using coordinate-based $\operatorname{\mathbb{C}MLP}$.
    (b) Comparison with SOTA learning-based image-to-image frameworks, TEMPO~\cite{ISPD-2020-TEMPO}, DOINN~\cite{DAC22-DOINN-Yang}.
  }
  \label{fig:flow}
\end{figure*}
\section{Framework}

This section covers the design flow of physics-informed Nitho,
including analysis of optical kernel dimensions,
details about optical kernel regression strategy using complex-valued neural fields,
and a new training paradigm for simulating forward lithography.

\subsection{Design of Predicted Optical Kernel Dimensions}
\label{subsec:tcc_dim}
For simplicity, the aerial intensity $\vec{I}$ in \Cref{eq:socs} can be written as,
\begin{equation}
  \vec{I}=\sum_{i}^{r} \left|\mathcal{F}^{-1}\left( \mathcal{K}_i  \odot \mathcal{F}(\boldsymbol{M})\right)\right|^2 ,
\end{equation}
where $\mathcal{K}_i$ is the $i$-th optical kernel, $r$ is the total number of kernels.

To design the predicted kernel dimensions, we find Mack introduced the \textit{theoretical resolution} in ~\cite{DFM-B2008-Mack}.
Let $R$ represents the resolution element (the line-width or the space-width) of mask patterns, the resolution is given by
$R = 0.5 * (\lambda / {N\!A})$. $\lambda$ is the exposure wavelength.
${N\!A}$ (\textit{numerical aperture}) is an important factor for the \textit{resolution limit} of the projection system,
which can be regarded as the ``\textit{collection efficiency}'' of the projector.
The smallest pitch that can be printed would put the first diffracted order at the largest angle that could pass through the lens,
defined by ${N\!A}$.
Using this description, we can set the kernel width and height as,
\begin{equation}
 m = (W \times \frac{2 \times {N\!A}}{\lambda}) \times 2 +1,~
 n = (H \times \frac{2 \times {N\!A}}{\lambda}) \times 2 +1,
 \label{eq:kernel_mn}
\end{equation}
where we use one-pixel width/height to represent 1$nm$,
the mask pitch can be replaced by mask image width $W$, height $H$.
Since the eigenvalues $\alpha_i$ in \Cref{eq:socs} rapidly decay in magnitude,
truncating the summation at order $r$ can be a decent approximation with error bounds proven in~\cite{Pati94}.
With discussions above, we can set a hyperparameter $r$ and design the optical kernel dimension as $\mathcal{K} \in \mathbb{C}^{r \times n \times m}$.
Given commonly used $\lambda\!=\!193 nm$, $N\!A\!=\!1.35$, we can obtain in \Cref{eq:kernel_mn}, $m\approx0.028*W$, $n\approx 0.028*H$.
Previous image learning-based models need to cover the prediction on $\mathbb{R}^{C\times W\times H}$ space while
Nitho is performing regression on a much smaller space $\mathbb{C}^{r\times n\times m}$ ($r < 60$ in our settings).
This can fundamentally contribute to our smaller model size.
Besides, we will show in subsequent experiments, due to the \textit{resolution limit},
even if $\lambda$ and $N\!A$ are unknown, the optical kernel dimensions can be obtained through a simple hyperparameter search.

\subsection{Complex-Valued Neural Fields for Optical Kernel Regression}
\label{subsec:cvnerf}
To predict the optical kernels, there are two critical issues to consider.
First, the optical kernels $\mathcal{K}$ are in a complex-valued formulation.
However, the vast majority of building blocks and architectures of previous
methods are based on real-valued operations and representations.
So we design a series of atomic complex-valued activation functions and layers
to assemble a complex-valued multilayer perceptron ($\operatorname{\mathbb{C}MLP}$) in \Cref{subsubsec:cvmlp}.
Second, the implicit optical kernel values need to be decoded from the mask-aerial image pairs.
The network's input should be carefully designed to eliminate the effect of mask layer types or data distribution.
Observing \Cref{eq:tcc}, TCC spectrum values in the spatial frequency domain are integrals over the frequency points $(f, g)$,
which means there will be a mapping between the spectrum coordinates and TCC optical kernel values.
Therefore, inspired by NeRF~\cite{ECCV-2020-NeRF},
we leverage ($\operatorname{\mathbb{C}MLP}$) in \Cref{subsubsec:nfokr} to perform optical kernel regression from coordinates.

\subsubsection{Complex-valued multilayer perceptron}
\label{subsubsec:cvmlp}
A complex number $z = a + ib$ has a real component $\Re(z)\!:=\!a$, and an imaginary component $\Im(z)\!:=\!b$.
The complex-valued neuron can be defined as $ o = \phi(\vec{x} \cdot \vec{w} + b)$, with an activation function $\phi$, applied to the input $\vec{x} \in \mathbb{C}^{n}$.
We can arrange complex neurons into a complex linear layer  ($\operatorname{\mathbb{C}Linear}$), $o = \phi(\vec{x}\vec{W} + b)$. $\vec{W}$ is layer weight.
\iffalse
In order to perform the equivalent of a traditional real-valued linear operation in the complex domain,
with which we can simulate complex arithmetic using real-valued entities and construct complex-valued linear operations.
We multiply a complex weight matrix $\vec{W} = \vec{A} + i\vec{B}$ by a complex vector $\vec{h} = \vec{x} + i\vec{y}$,
where $\vec{A}$ and $\vec{B}$ are real matrices and $\vec{x}$ and $\vec{y}$ are real vectors. We obtain:
\begin{equation}
  \vec{W} \vec{h}  = (\vec{A} \vec{x} - \vec{B} \vec{y}) + i(\vec{B}\vec{x} + \vec{A}\vec{y}),
\end{equation}
\fi
The differentiability of complex-valued linear operations can be proved in the paper~\cite{DeepCplxNets}.
Complex rectified linear unit ($\operatorname{\mathbb{C}ReLU}$) is applied as the activation function:
\begin{equation}
  \operatorname{\mathbb{C}ReLU}(z)=\operatorname{ReLU}(\Re(z))+i \operatorname{ReLU}(\Im(z)).
\end{equation}
As illustrated in \Cref{fig:nitho}, the $\operatorname{\mathbb{C}MLP}$ is further constructed as,
\begin{equation}
  \operatorname{\mathbb{C}MLP}: \operatorname{\mathbb{C}Linear}\!\to\!(\operatorname{\mathbb{C}Linear}\!\to\!\operatorname{\mathbb{C}ReLU})\!\times\!N\!\ldots\!\to\!\operatorname{\mathbb{C}Linear},
\end{equation}
where $\times N$ means there are $N$ hidden blocks $(\operatorname{\mathbb{C}Linear}\!\to\!\operatorname{\mathbb{C}ReLU})$.

\subsubsection{Neural fields for optical kernel regression}

% DNF: Appendix A
A recent trend in computer vision and graphics research is replacing discrete representations with
\textit{coordinate-based neural representations}.
Given the recent success of view synthesis,
NeRF~\cite{ECCV-2020-NeRF} presents a continuous scene as a 5D vector-valued function whose input is a 3D location $\vec{x}\!=\!(x, y, z)$ and viewing direction $(\theta, \phi)$,
and whose output is an emitted color $\vec{c}\!=\!(r, g, b)$ and volume density $\sigma$ at that location.
Then NeRF approximates the continuous 5D scene representation with an MLP network $F_{\Theta}:(\mathbf{x}, \mathbf{d})\!\to\!(\mathbf{c}, \sigma)$
and optimizes its weights $\Theta$ to map from each input 5D coordinate to its corresponding volume density and directional emitted color.

\label{subsubsec:nfokr}
Considering the industrial case, as drawn in \Cref{fig:lens},
the lithographic imaging system consists of a source, a pupil, and a set of lenses, independent of masks.
As mentioned, imaging functions are location-dependent on the wafer plane,
meaning there is an implicit mapping between the optical kernel coordinates and values.
Similar to NeRF, Nitho takes the 2D coordinates $(x, y)$ in optical kernel dimension space $\mathbb{R}^{n \times m}$,
a constant matrix independent of mask types or data distribution,
as the inputs of our $\operatorname{\mathbb{C}MLP}$.
Then leverage the $\operatorname{\mathbb{C}MLP}$ as the implicit mapping function to best recover the complex-valued optical kernels $\mathcal{K}$ from coordinates:
\begin{equation}
  \hat{\mathcal{K}} = \operatorname{\mathbb{C}MLP}(\vec{v}; \Theta) .
\end{equation}
Each of input coordinate $\vec{v}_i=(x,y)$ specify the index of diffraction orders of mask / TCC spectrum.
And the outputs of neural fields stand for the predicted optical kernels $\hat{\mathcal{K}}$, \ie TCC spectrum matrix, as depicted in \Cref{fig:nitho}.
The predicted aerial image $\hat{\vec{I}}$ can be obtained using SOCS formula in \Cref{eq:socs}.
Physically, each predicted kernel $\hat{\mathcal{K}}_i$ represents \textit{a coherent point on the source}~\cite{DFM-B2008-Mack}.
Using coordinates as input also ensures a fair comparison with previous works,
since no extra information is needed beyond mask-aerial pairs.
Like the real simulator,
the separation of kernel regression and mask processing eliminates the dependence on data distribution to the greatest extent possible,
thus improving generalization and accuracy.

\subsubsection{Positional encoding}
To compensate for $\operatorname{MLP}$'s inability to represent high-frequency signals,
NeRF uses \textit{positional encoding} (PE) to map the coordinates from $\mathbb{R}$ to a higher dimensional space $\mathbb{R}^{2L}$:
\begin{equation}
  \small
  \gamma(\vec{v})\!=\!\left[\sin (2^{0}\pi\vec{v}),\!\cos (2^{0}\pi\vec{v}),\!\ldots,\!\sin\!\left(2^{L-1}\pi \vec{v}\!\right),\!\cos\!\left(2^{L-1}\pi\vec{v}\!\right)\right]^{\mathrm{T}}\!,
  \label{eq:nerf_pe}
\end{equation}
here $\gamma(\cdot)$ is a positional mapping applied separately on each of the two coordinate values in $\vec{v}$, $L$ is a hyperparameter.
The fidelity of NeRF depends critically on the positional encoding,
as it allows $\operatorname{MLP}$ to parameterize both low-frequency and high-frequency embeddings.
However, Nitho confronts varied challenges.
Directly utilizing of the NeRF's PE will cause a failure.
The PE of NeRF in \Cref{eq:nerf_pe} employs solely on-axis frequencies, which is ideal for NeRF's ray rendering tasks.
Nonetheless, due to the absence of a strong prior of optical kernel frequency distribution,
the axis-aligned mapping used in NeRF will decrease the performance of Nitho.
Therefore, we adopt Gaussian random Fourier feature (RFF)~\cite{tancik2020rff} mapping, which has an isotropic frequency distribution,
as our positional encoding:
\begin{equation}
  \gamma(\vec{v})=[\cos (2 \pi \vec{B v})*(1+j),~\sin (2 \pi \vec{B v})*(1+j)]^{\mathrm{T}},
  \label{eq:cplx_gaussian_pe}
\end{equation}
where each entry in $\vec{B} \in \mathbb{R}^{l\times d}$ is sampled from $\mathcal{N}(0, \sigma^2)$, and $\sigma$ is a hyperparameter for standard deviation.
Coordinates $\vec{v}$ will be normalized to lie in $[0, 1]$, and each $\sin$ and $\cos$ entry will be multiplied by $(1+j)$ to map the coordinates to complex-valued fields.
Now the optical kernels are predicted as:
\begin{equation}
  \hat{\mathcal{K}} = \operatorname{\mathbb{C}MLP}(\gamma(\vec{v}); \Theta) .
\end{equation}

\subsection{Forward Training Procedure of Nitho}
\label{subsec:forward_training}

\begin{algorithm}[h]
  \caption{\fname~Forward Training Procedure}
    \begin{algorithmic}[1]
        \Require Mask image set $\mathcal{M}_{tr}$, aerial image set $\mathcal{I}_{tr}$.
        \Ensure Predicted TCC optical kernels  $\mathcal{\hat{K}}$
        \State $m, n \gets$ optical kernel width, height using \Cref{eq:kernel_mn} \label{ag1:line:get_mn}
        \State $\vec{v} \gets$ coordinate space of $\mathbb{R}^{n\times m}$ \label{ag1:line:get_xy}
        \State $\vec{v}_p \gets \gamma(\vec{v})$ \Comment{Complex-valued positional encoding.}
        \ForAll{$\vec{M} \in \mathcal{M}_{tr}$} \label{ag1:line:start_training}
        \State $\vec{I} \in \mathcal{I}_tr \gets$ the aerial image ground truth of $\vec{M}$ \label{ag1:line:get_aerial}
        \State $\mathcal{F}(\vec{M}) \gets \texttt{fftshift}(\texttt{fft2}(\vec{M}))$ \label{ag1:line:fftm}
        \State $\mathcal{F}(\vec{M}) \gets $ Crop $\mathcal{F}(\vec{M})$ centrally to width $m$, height $n$ \label{ag1:line:fftm_crop}
        \State $\hat{\mathcal{K}} = \operatorname{\mathbb{C}MLP}(\vec{v}_p ; \Theta)$  \Comment{Predicting kernels $\hat{\mathcal{K}}$.} \label{ag1:line:get_tcc}
        \State $r \gets$ order number of predicted kernel $\hat{\mathcal{K}} \in \mathbb{C}^{r \times n \times m}$ \label{ag1:line:order_n}
        \For{$ i \in \{1, \ldots r\}$ } \Comment{SOCS formula.} \label{ag1:line:socs_start}
        \State $ \vec{E}_i = \mathcal{F}^{-1}(\hat{\mathcal{K}}_i * \mathcal{F}(\vec{M}))$ \Comment{Electric field $\vec{E}_i$} \label{ag1:line:ei}
        \State $\hat{\vec{I}} = \hat{\vec{I}} + |\vec{E}_i * \vec{E}_i^{*}|$  \Comment{Converting to intensity} \label{ag1:line:socs_end}
        \EndFor
        \State $Loss = \operatorname{MSE}(\vec{I}, \hat{\vec{I}})$  \Comment{Predicted aerial image $\hat{\vec{I}}$} \label{ag1:line:mse}
        \EndFor
    \end{algorithmic}
  \label{alg:Nitho}
\end{algorithm}

\Cref{alg:Nitho} shows the forward training procedure of Nitho framework, also illustrated in \Cref{fig:nitho}.
In \cref{ag1:line:get_mn}, we calculate the kernel width/height based on \Cref{eq:kernel_mn} to set up dimensions of learning target $\hat{\mathcal{K}}$.
Next, in \cref{ag1:line:get_xy}, the kernel coordinates are flattened as a matrix: $[(0, 0), \ldots, (0, m), \dots, (n, m)]^{\mathrm{T}}$,
then get positional encoding in a higher dimension of complex fields by \Cref{eq:cplx_gaussian_pe}.
The training process starts from \cref{ag1:line:start_training}.
We select training pairs $\vec{M}$ and $\vec{I}$, and get mask spectrum $\mathcal{F}(\vec{M})$ by 2D-FFT with necessary shifting.
Then crop the central region of mask spectrum to match the kernel dimension. (\Cref{ag1:line:get_aerial,ag1:line:fftm,ag1:line:fftm_crop}).
In \cref{ag1:line:get_tcc}, the positional encoding of kernel coordinates $\gamma(\vec{v})$ is provided as input to a $\operatorname{\mathbb{C}MLP}$
parameterized by complex-valued weight $\Theta$, whose outputs are the predicted optical kernels $\hat{\mathcal{K}}$.
Finally, we can get the predicted aerial image $\hat{\vec{I}}$ using SOCS formula in \Cref{eq:socs} (\cref{ag1:line:socs_start,ag1:line:ei,ag1:line:socs_end}).
Note that $\vec{E}_i^{*}$ in \cref{ag1:line:socs_end} is the complex conjugate of $\vec{E}_i$, thus $\hat{\vec{I}}$ is a real-valued matrix.
We can apply $\operatorname{MSE}$ loss on predicted aerial image $\hat{\vec{I}}$ and its ground truth $\vec{I}$ (\cref{ag1:line:mse}).
As FFT is a differentiable linear operation, the whole Nitho is differentiable. The weight $\Theta$ can be optimized by gradient descent.
\subsubsection{Fast lithography}
Nitho also significantly accelerates forward lithography.
Different from previous work, Nitho has no network inference process after training.
The predicted optical kernels can be stored in the same manner as the actual TCC kernels to perform SOCS calculation (\cref{ag1:line:get_aerial,ag1:line:fftm,ag1:line:fftm_crop,ag1:line:get_tcc,ag1:line:order_n,ag1:line:socs_start,ag1:line:ei,ag1:line:socs_end}).
We also propose a hierarchical GPU acceleration strategy for SOCS.
In \cref{ag1:line:fftm}, the FFT is operated on GPU using the well-optimized \texttt{PyTorch} FFT library.
And the summation in \cref{ag1:line:socs_start} can be performed parallelly to get the final aerial image.

\subsection{Comparisons between Nitho and SOTA}

\begin{table}[htb!]
	\centering
	\caption{Comparisons between Nitho and SOTA.}
	\label{tab:diff}
	\begin{tabular}{l|ccccc}
		\toprule
		                  & TEMPO~\cite{ISPD-2020-TEMPO}  & DOINN~\cite{DAC22-DOINN-Yang}  & Nitho            \\ \midrule
    Training pair     & Mask-Aerial                  & Mask-Resist                    & Mask-Aerial        \\
		Network Modeling  & $S(\mathcal{T} * G(\cdot))$ & $H(S(\mathcal{T} * G(\cdot)))$ & $\mathcal{F}(\mathcal{T})$        \\
	  Network Arch.    & cGAN                        & FNO+CNN          & $\operatorname{\mathbb{C}MLP}$    \\
	  Network Size      & $\sim$ 31 MB              & $\sim$1.3 MB          &  0.41 MB   \\ \bottomrule
	\end{tabular}
\end{table}

To compare Nitho and SOTA with greater clarity,
the lithography model can be reduced to:
\begin{equation}
  \vec{I} = S(\mathcal{T} * G(\vec{M})), ~ \vec{Z} = H(\vec{I} - I_{thres}),
\end{equation}
where $S$, $G$ are mask-related linear operations.
$H$ is an image binarization function.
$\mathcal{T}$ is a constant optical matrix.
In \Cref{tab:diff}, we list two previous works TEMPO~\cite{ISPD-2020-TEMPO} and DOINN~\cite{DAC22-DOINN-Yang},
which are the SOTA of the aerial image stage and resist image stage, respectively.
As illustrated in \Cref{fig:prev},
TEMPO uses cGAN to model the mask-to-aerial process $S(\mathcal{T} * G(\cdot))$.
DOINN learns the mask-to-resist process $H(S(\mathcal{T} * G(\cdot)))$.
Differently, Nitho models the mask independent optical kernels $\mathcal{F}(\mathcal{T})$, which contributes to superior generalizability with a smaller model size.
\section{Experiment}

\subsection{Datasets}
In \Cref{tab:data}, our model is evaluated on both metal and via layer designs.
The t-SNE distribution of datasets is shown in \Cref{fig:general}(a).
\subsubsection{ICCAD-2013}
ICCAD 2013 CAD contest~\cite{OPC-ICCAD2013-Banerjee} provides ten 4$\mu m^2$ tiles for testing.
We obtain the training set from GAN-OPC~\cite{OPC-TCAD2020-Yang},
which contains 4K 4$\mu m^2$ tiles generated following the same design rules of the contest.
To test the robustness of the model, we also generate OPC'ed mask using MOSAIC~\cite{OPC-DAC2014-Gao}.
We apply the lithography simulator from~\cite{OPC-ICCAD2013-Banerjee} to obtain the golden aerial and resist images.
Different from previous art DOINN~\cite{DAC22-DOINN-Yang}, we directly use the highest resolution,
\ie 4$\mu m^2$ tiles are converted to $2000 \times 2000$-pixels images.

\subsubsection{ISPD-2019}
ISPD 2019 initial detailed routing contest~\cite{ispd2019-benchmark} provides designs synthesized with commercial placement and routing tools.
We randomly choose 4$\mu m^2$ tiles from ISPD-2019 designs for a fair comparison.
Unlike DOINN only tests its model on ISPD via layers, we choose both via layers and metal layers.
\begin{itemize}
  \item ISPD-2019 via layers: we apply the high-resolution settings from DOINN, 10K training set and 10K testing set with  4$\mu m^2$ tiles converted to
  $2000 \times 2000$ images.
  \item ISPD-2019 metal layers: we randomly select 1K metal patterns for training and 300 testing patterns.
  All metal patterns will be cropped into 4$\mu m^2$ tiles and converted to $2000\times 2000$ images.
\end{itemize}
The ground truth for ISPD metal and via layers are generated by commercial tool Mentor Calibre~\cite{TOOL-calibre} with $\lambda = 193 nm, N\!A=1.35$.

\begin{table}[tb!]
	\centering
	\caption{Details of the Dataset.}
	\label{tab:data}
	\setlength{\tabcolsep}{3pt}
	\renewcommand{\arraystretch}{.9}
	\begin{tabular}{l|l|cccc}
		\toprule
		\multicolumn{1}{c|}{Dataset}     & Alias &Train  & Test  & Tile Size  & Litho Engine \\ \midrule
		ICCAD-2013  & B1      & 4875  & 10    & 4$\mu m^2$ & Lithosim \cite{OPC-ICCAD2013-Banerjee}  \\
	  ICCAD-2013 (OPC) & B1opc & -     & 10    & 4$\mu m^2$ & Lithosim \cite{OPC-ICCAD2013-Banerjee}  \\
		ISPD-2019-metal & B2m     & 1000  & 300 & 4$\mu m^2$ & Calibre \cite{TOOL-calibre}  \\
		ISPD-2019-via   & B2v     & 10000 & 10000 & 4$\mu m^2$ & Calibre \cite{TOOL-calibre}  \\ \bottomrule
	\end{tabular}
\end{table}

\subsection{Results Comparison with State-of-the-Art}
\subsubsection{Model performance}

\begin{table}[tb!]
  \centering
  \setlength{\tabcolsep}{.5pt}
  \renewcommand{\arraystretch}{.5}
  \begin{tabular}{lcccccc}
      \midrule
      (a) B1   & \includegraphics[width=.135\linewidth,valign=m]{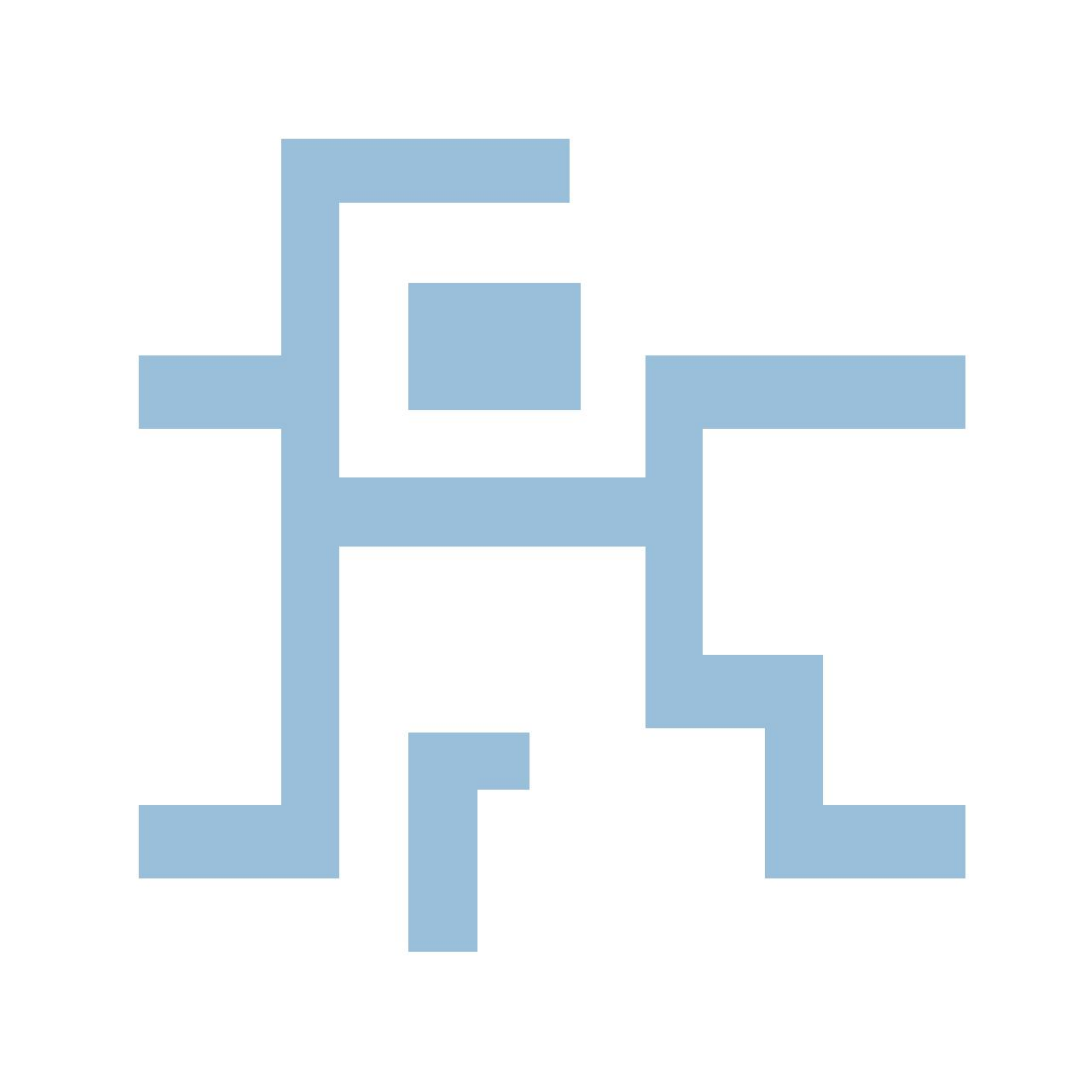}   & \includegraphics[width=.135\linewidth,valign=m]{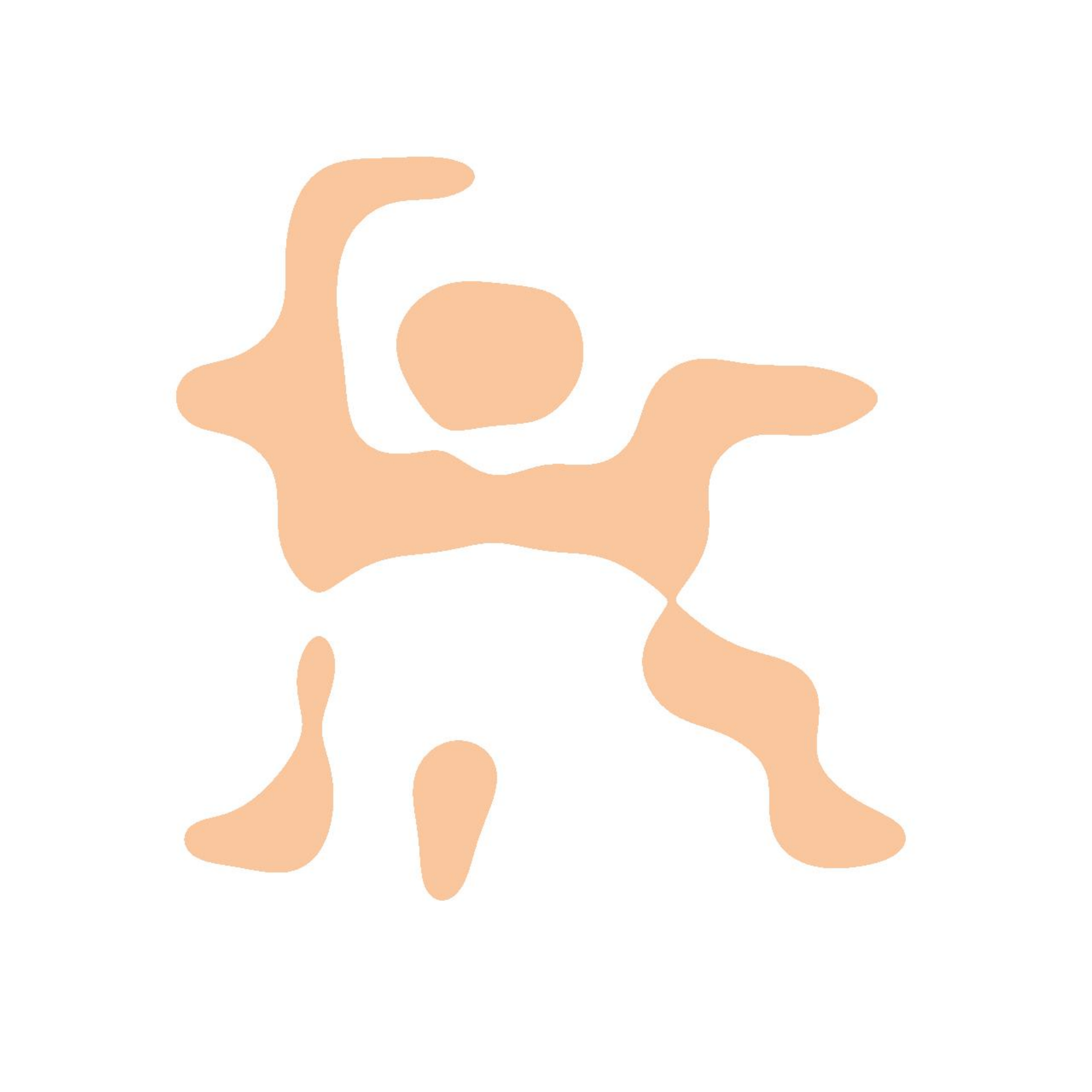}  & \includegraphics[width=.135\linewidth,valign=m]{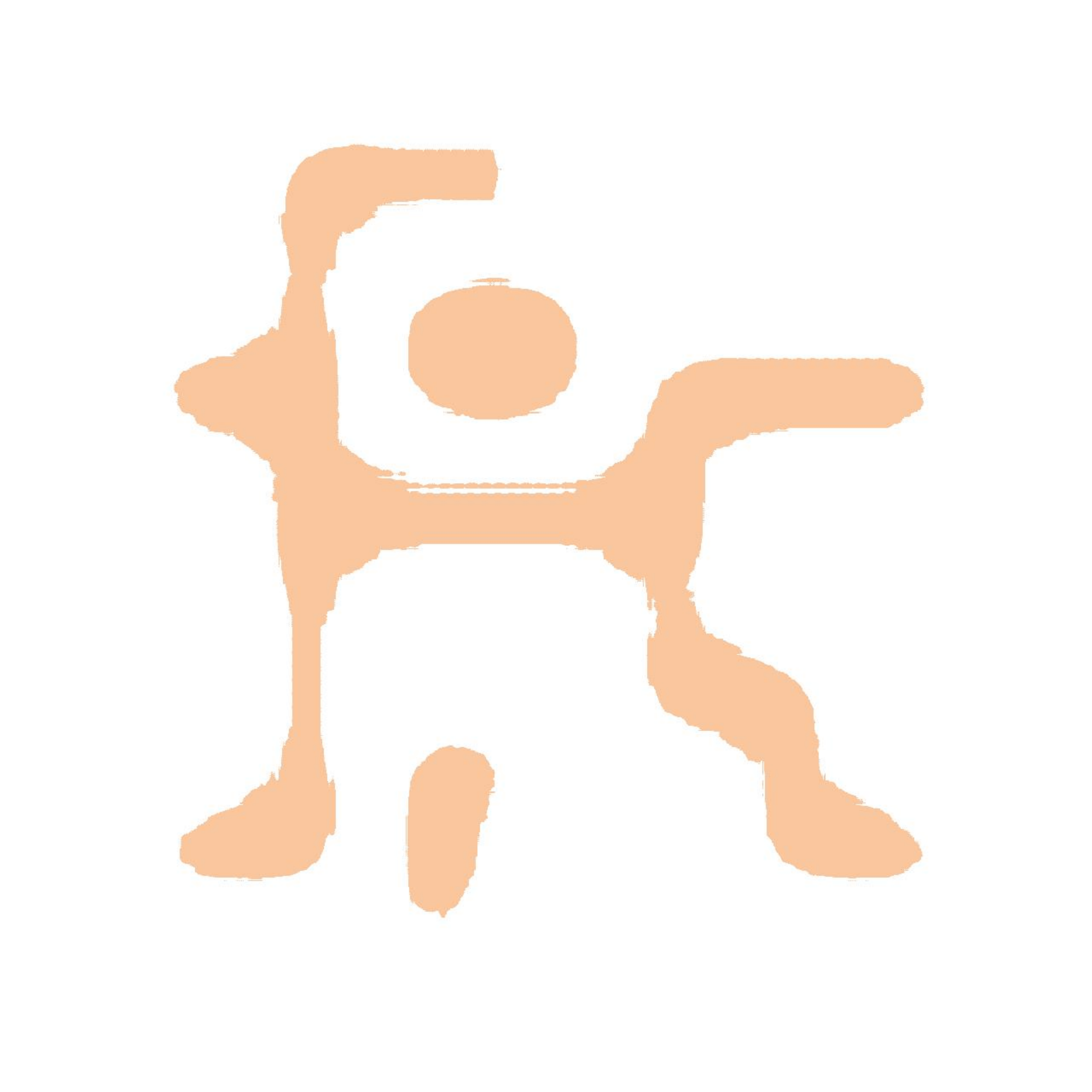} & \includegraphics[width=.135\linewidth,valign=m]{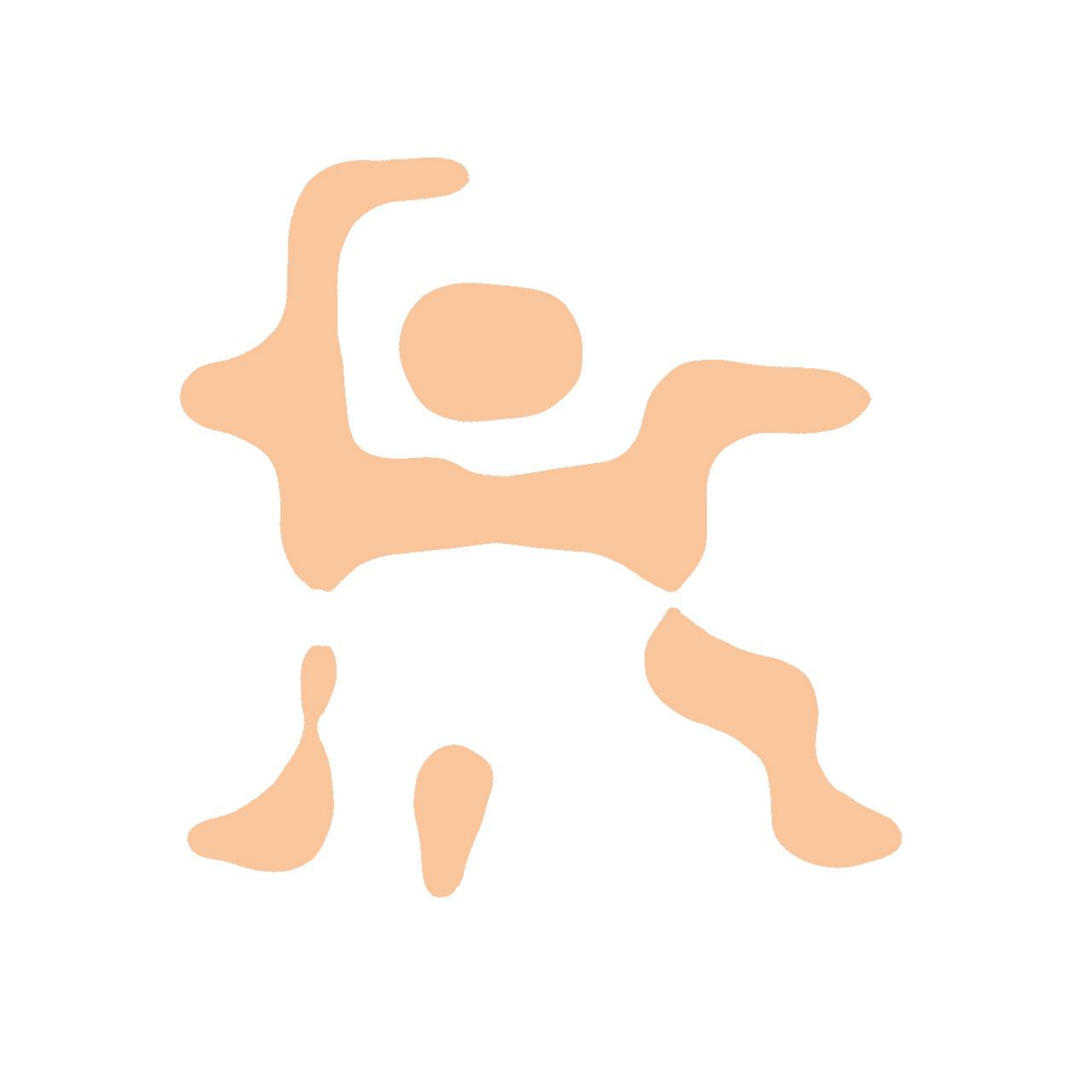}  & \includegraphics[width=.135\linewidth,valign=m]{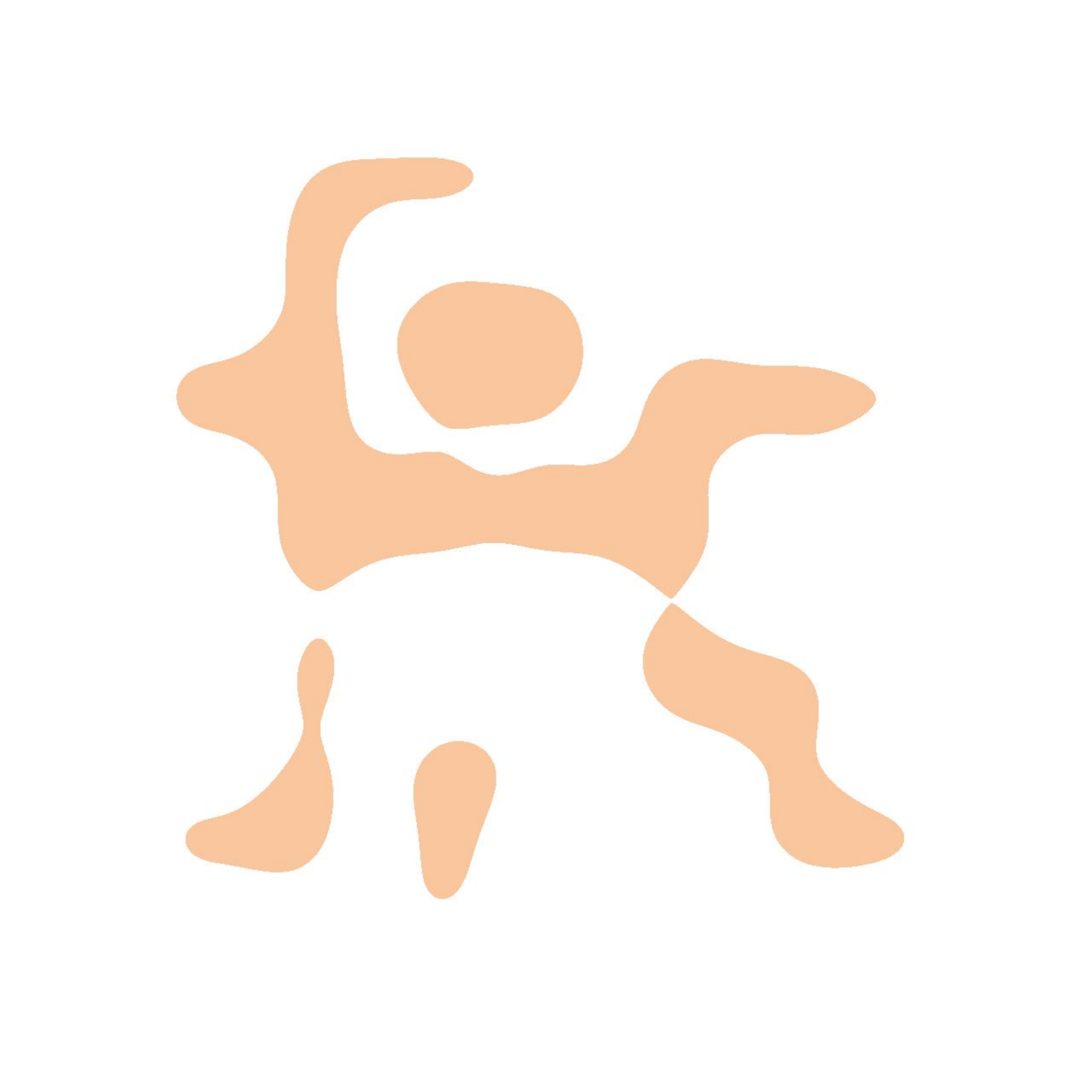} & \includegraphics[width=.135\linewidth,valign=m]{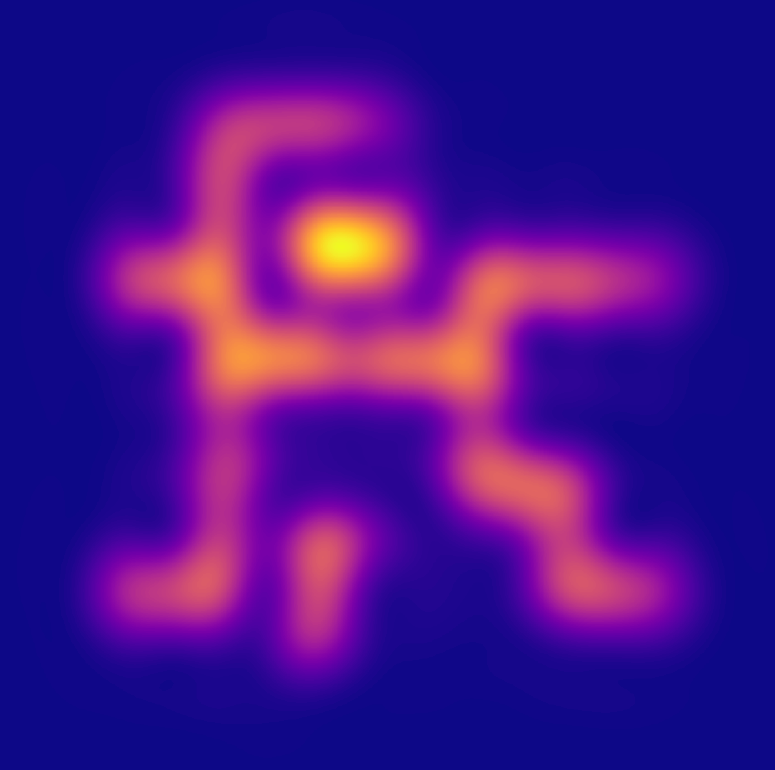} \\ \midrule
      (b) B2m  & \includegraphics[width=.135\linewidth,valign=m]{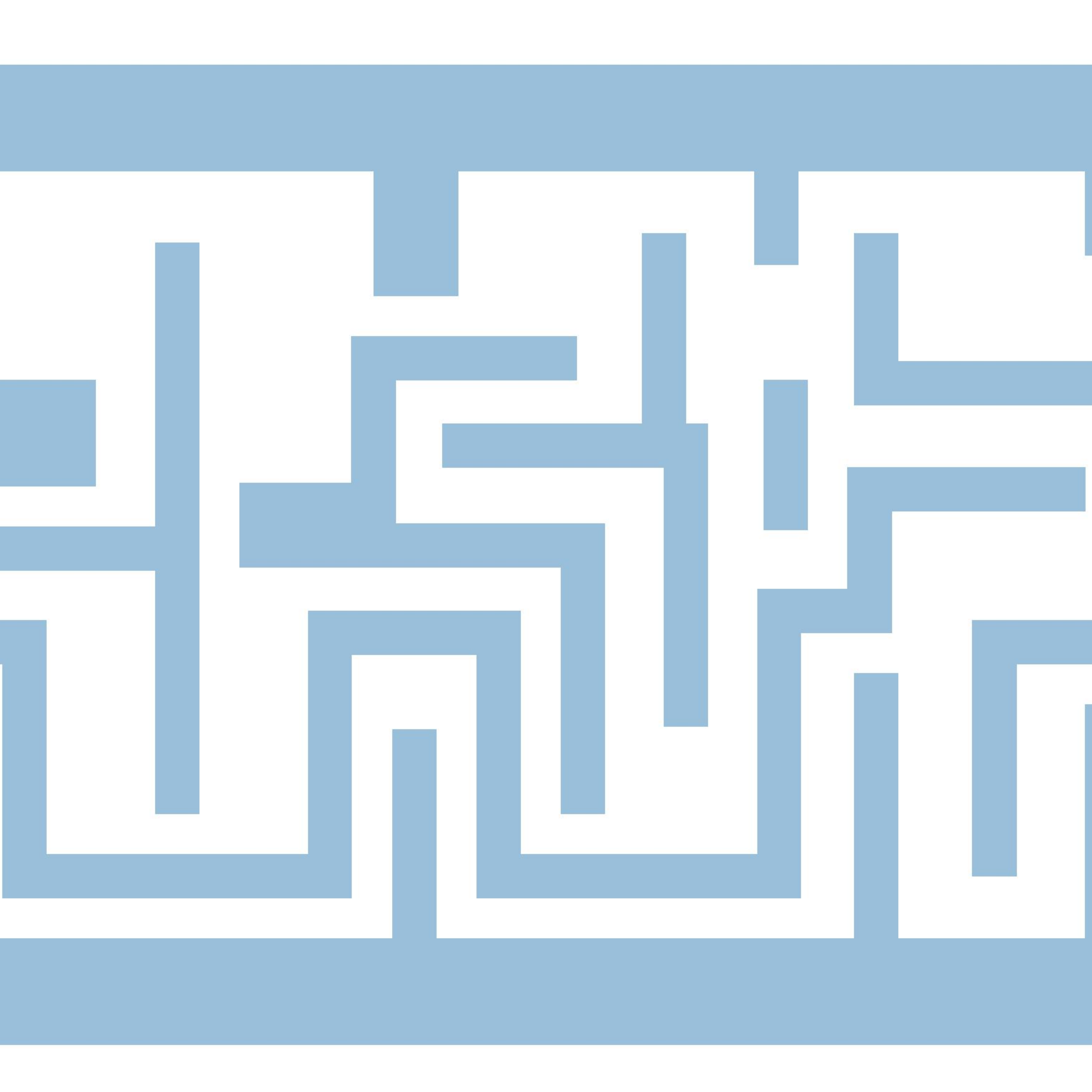}   & \includegraphics[width=.135\linewidth,valign=m]{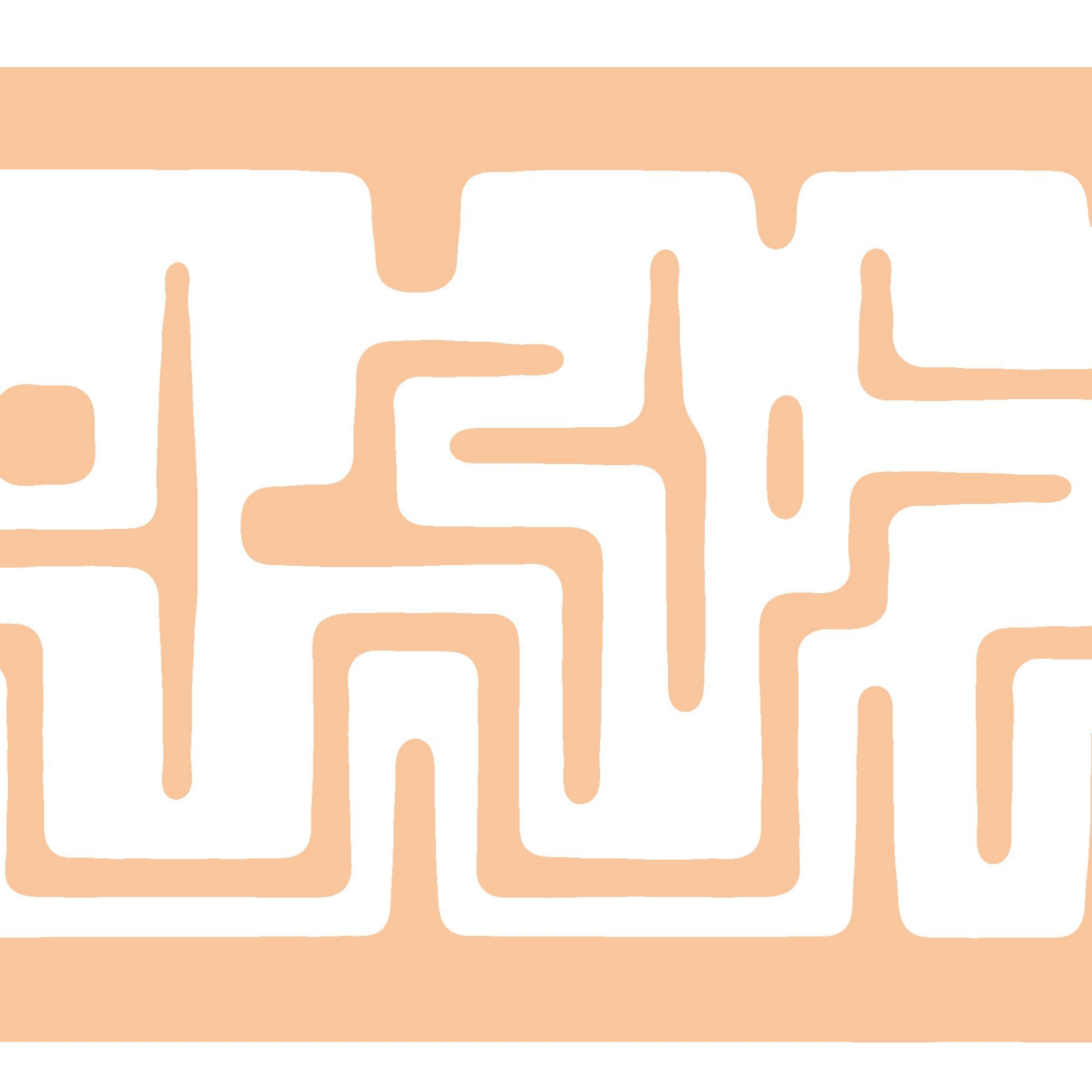}  & \includegraphics[width=.135\linewidth,valign=m]{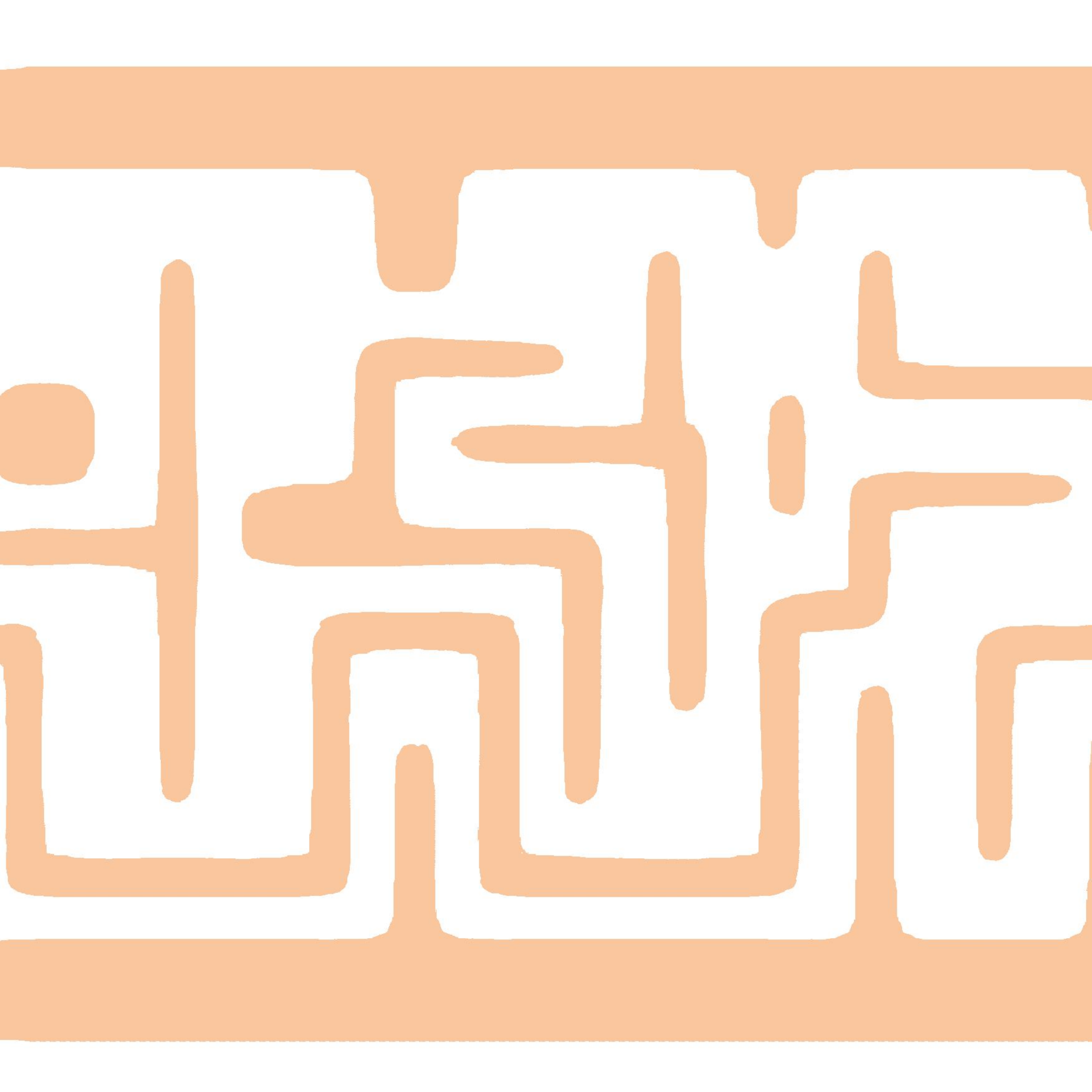} & \includegraphics[width=.135\linewidth,valign=m]{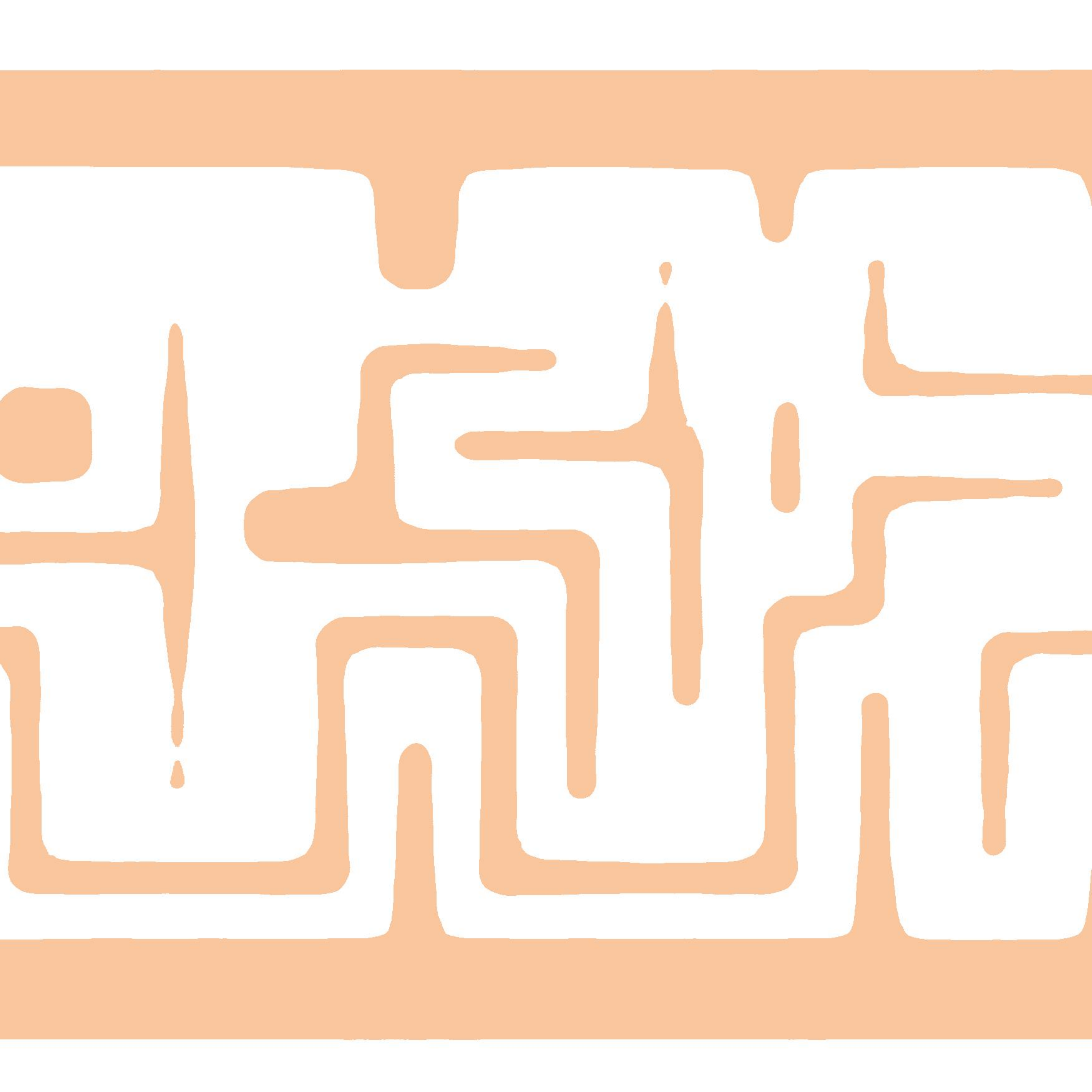}  & \includegraphics[width=.135\linewidth,valign=m]{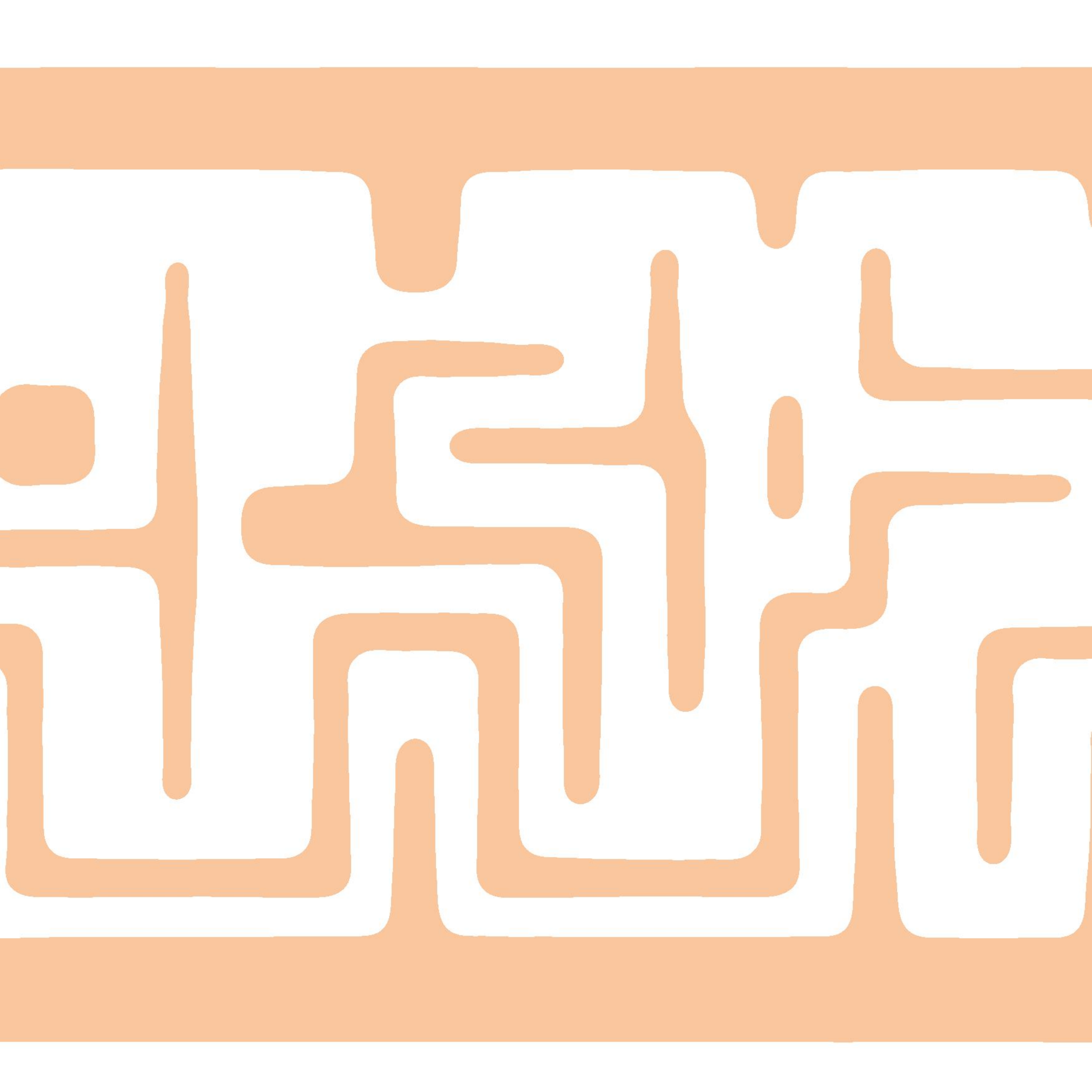} & \includegraphics[width=.135\linewidth,valign=m]{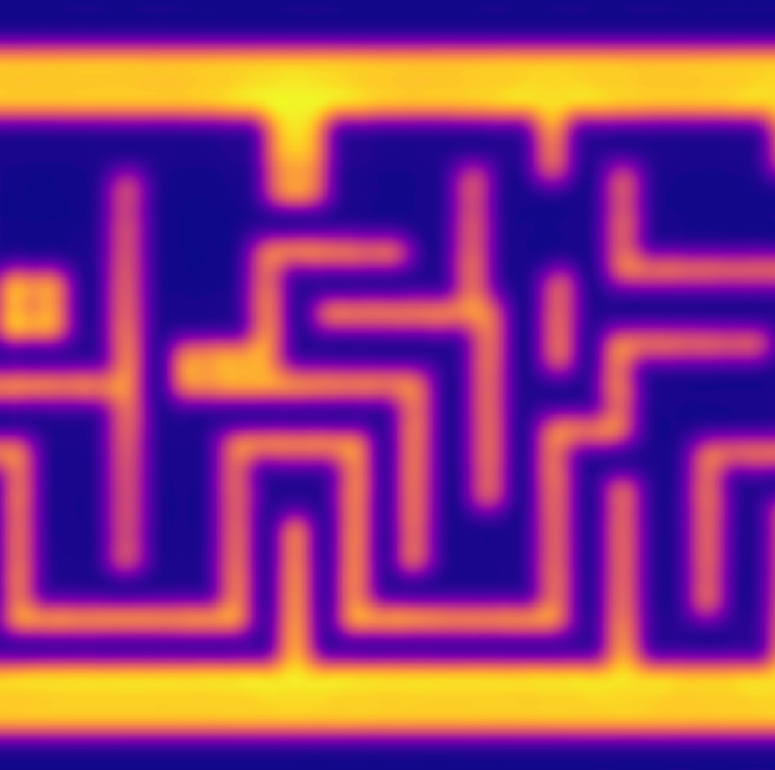}\\ \midrule
      (c) B2v  & \includegraphics[width=.135\linewidth,valign=m]{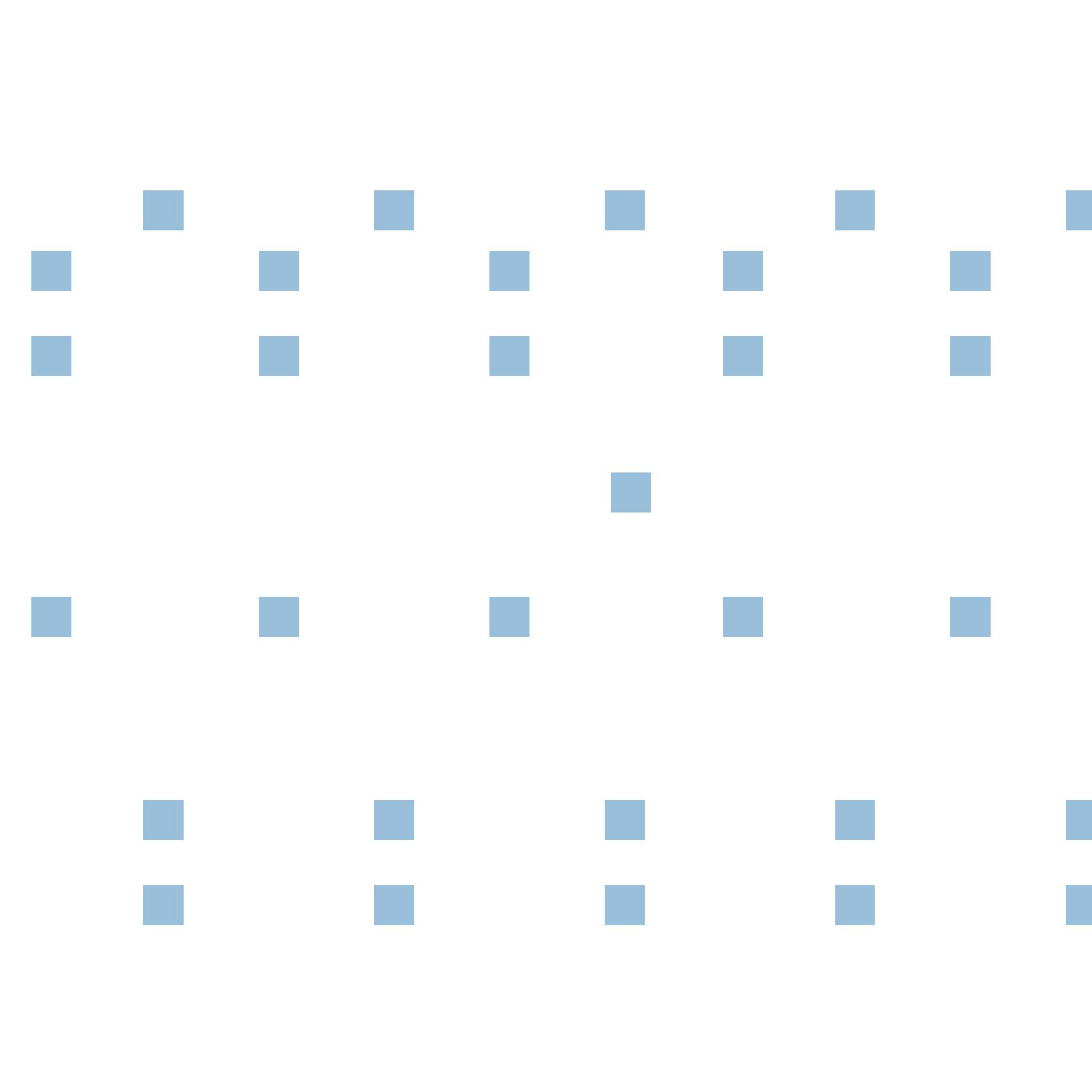}   & \includegraphics[width=.135\linewidth,valign=m]{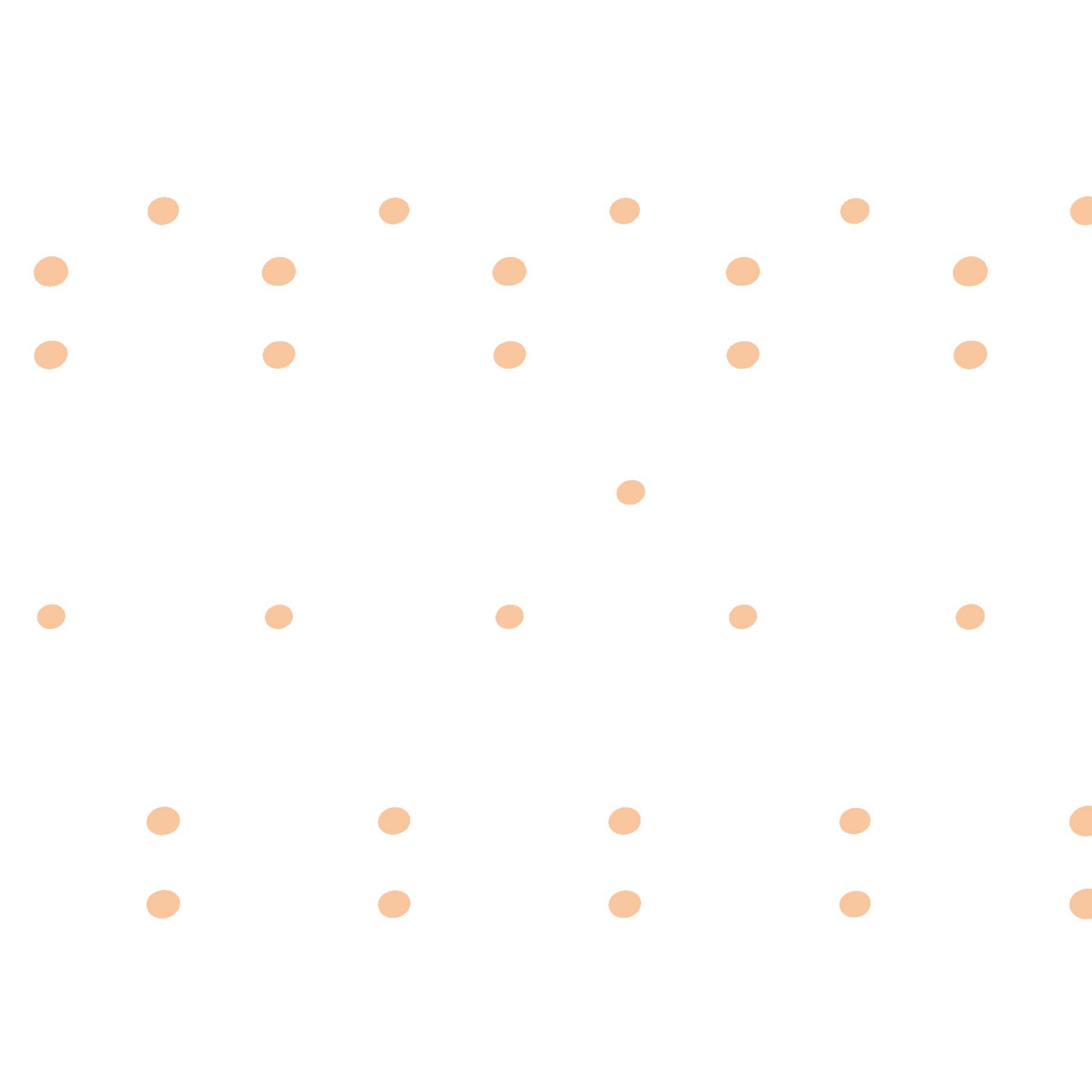}  & \includegraphics[width=.135\linewidth,valign=m]{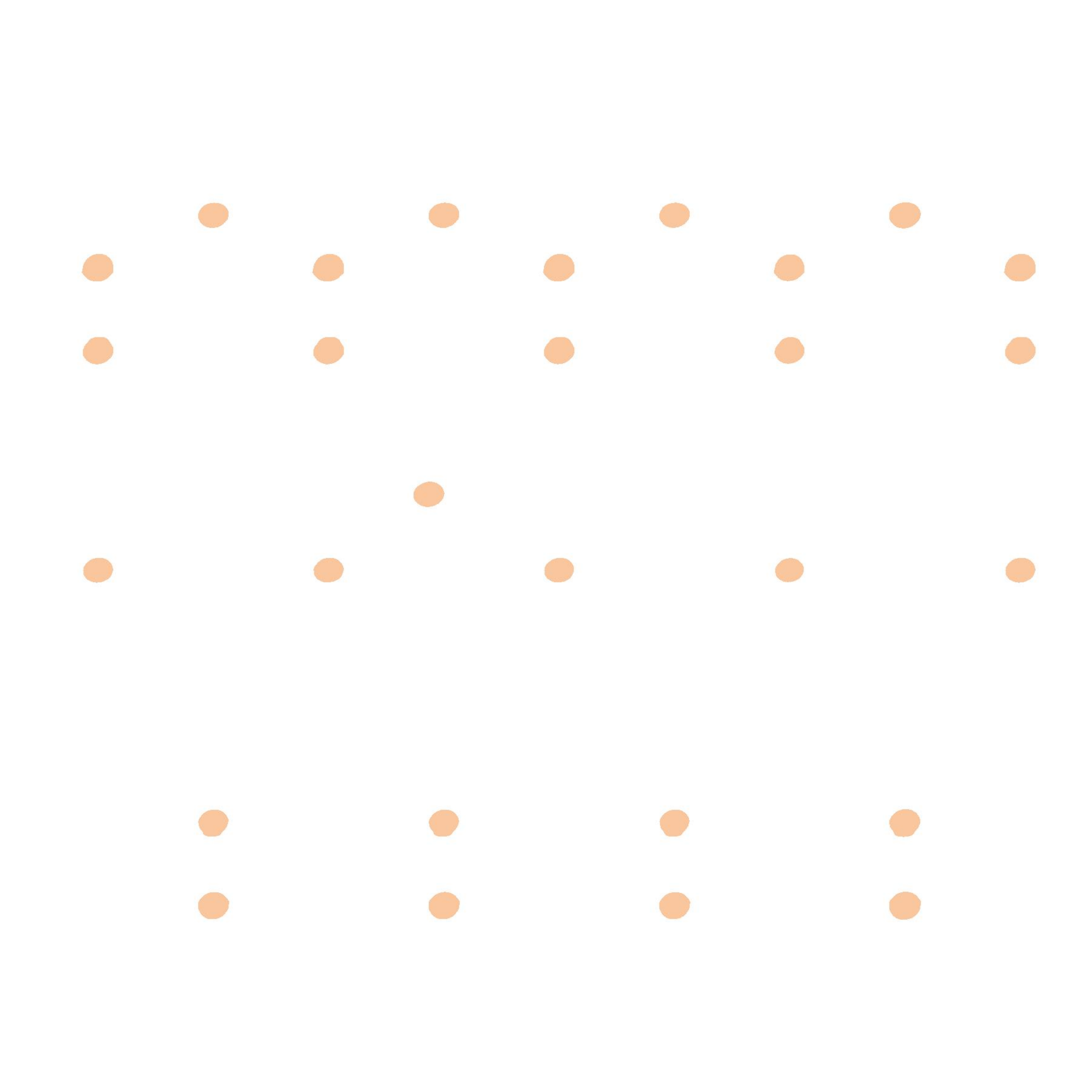} & \includegraphics[width=.135\linewidth,valign=m]{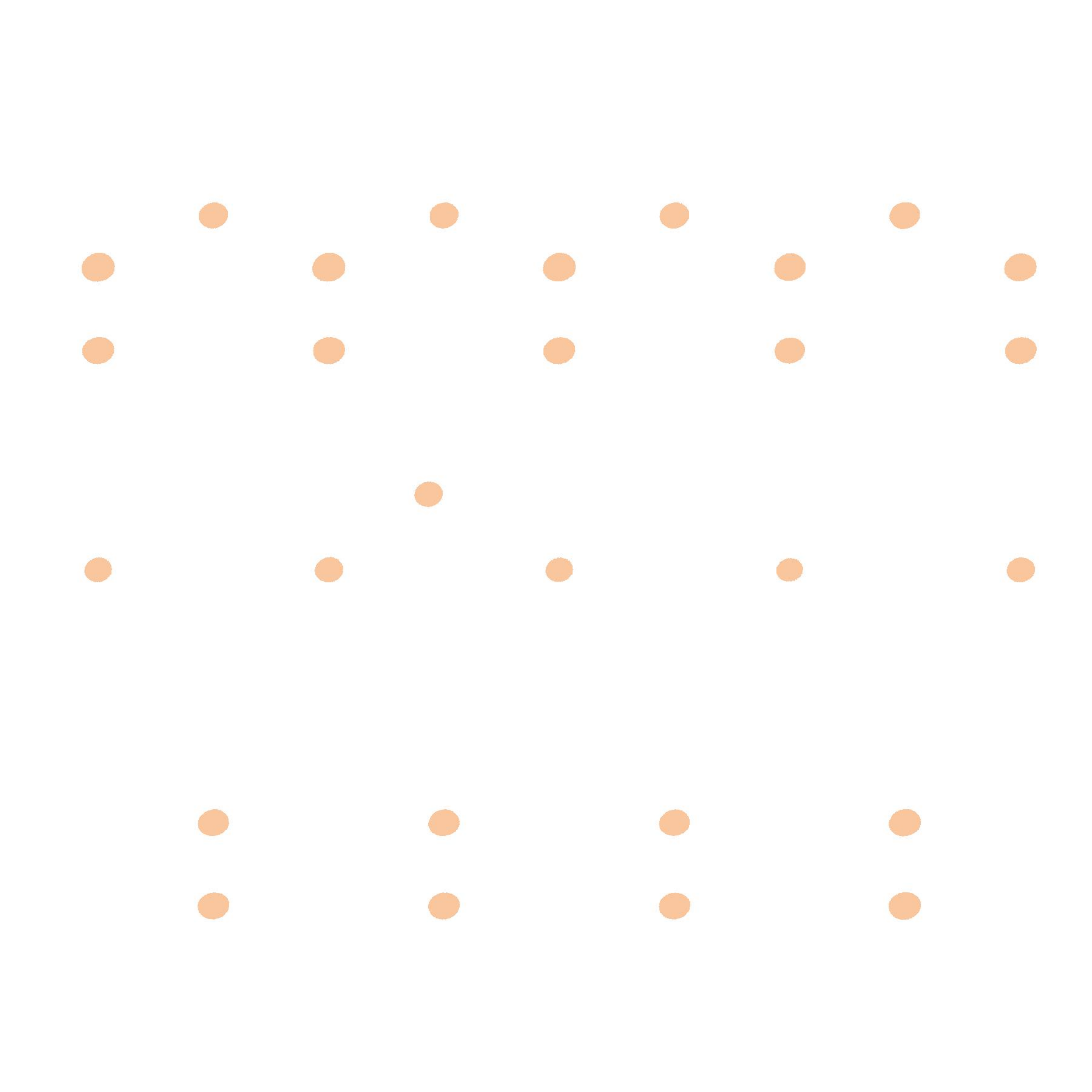}  & \includegraphics[width=.135\linewidth,valign=m]{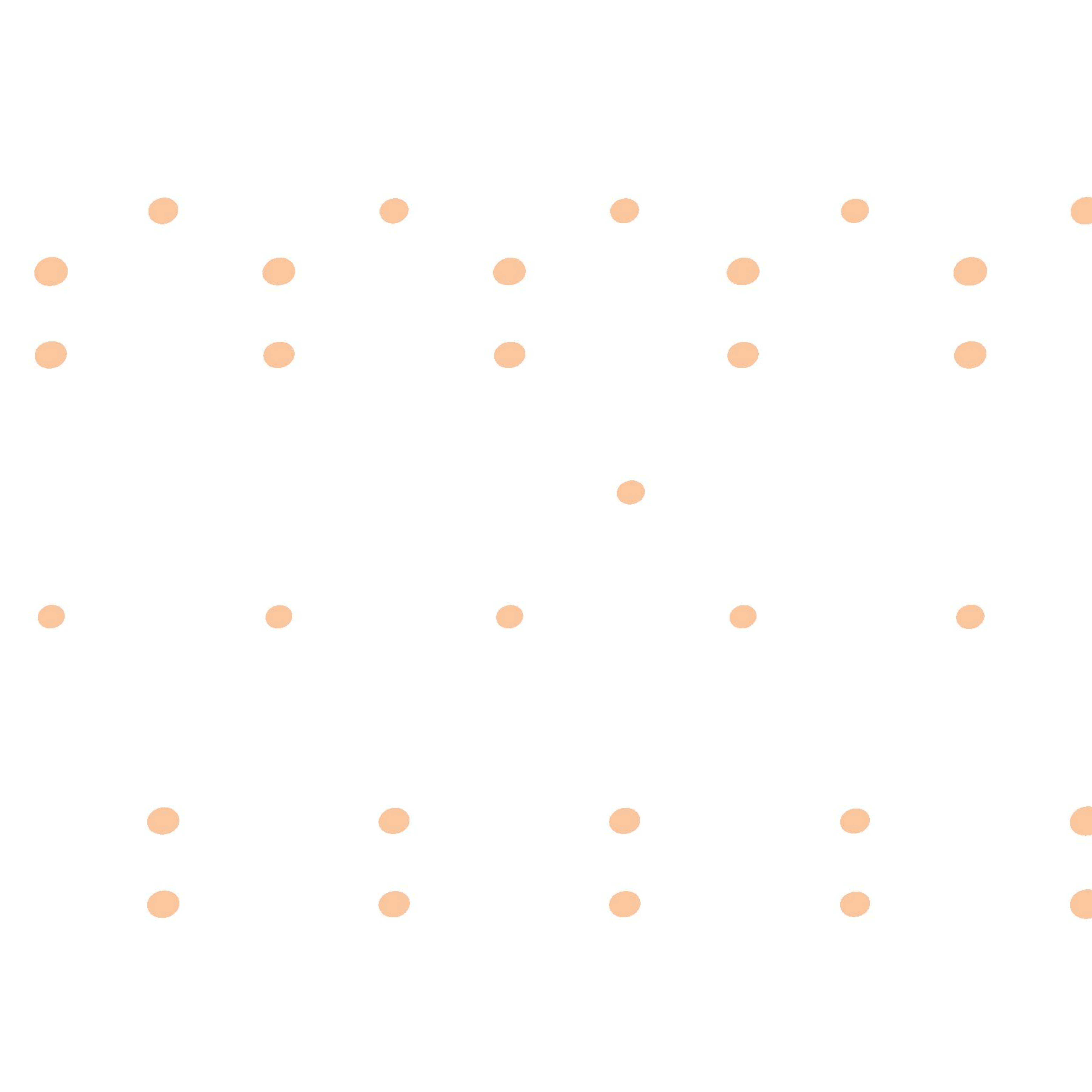}  & \includegraphics[width=.135\linewidth,valign=m]{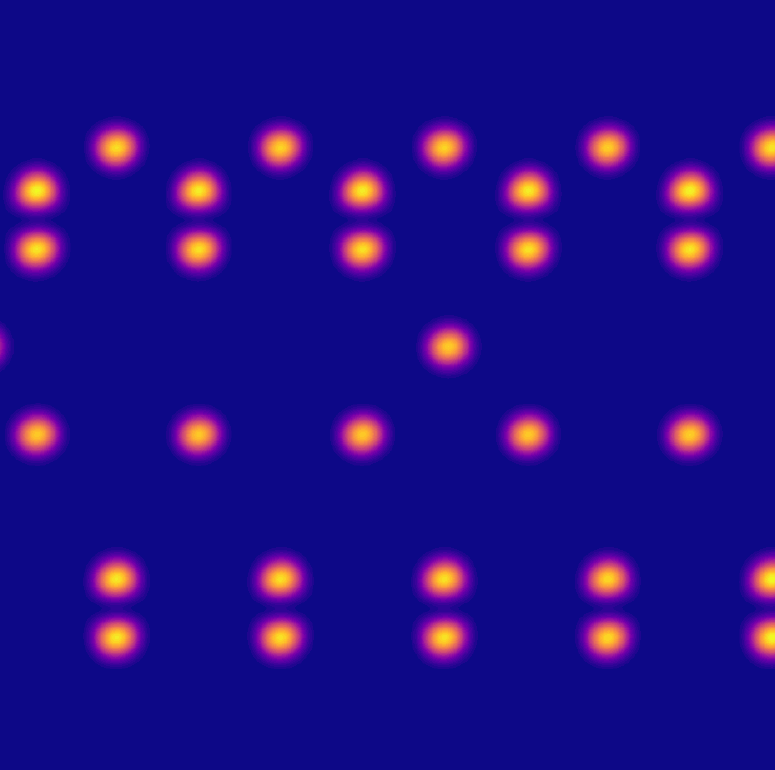}\\ \midrule
      & Mask   & Resist G.T. &  TEMPO  &  DOINN & Ours &{\footnotesize Our Aerial} \\
  \end{tabular}
  \captionof{figure}{Visualization of the results of Nitho in aerial and resist stage.}
  \label{fig:vis_masks}
\end{table}
\begin{table*}[htbp]
  \centering
  \caption{Result Comparison with State-of-the-Art.}
  \setlength{\tabcolsep}{5pt}
	\renewcommand{\arraystretch}{1}
  \makebox[\textwidth][c]{
    \begin{tabular}{c|ccc|ccc|ccc|cc|cc|cc}
      \toprule
      \multirow{4}{*}{Bench} & \multicolumn{9}{c|}{Aerial Image}                                                                                            & \multicolumn{6}{c}{Resist Image}                                                   \\
      \cmidrule(lr){2-10} \cmidrule(lr){11-16}
                             & \multicolumn{3}{c|}{TEMPO~\cite{ISPD-2020-TEMPO}}                & \multicolumn{3}{c|}{DOINN~\cite{DAC22-DOINN-Yang}}               & \multicolumn{3}{c|}{Nitho}             & \multicolumn{2}{c|}{$\text{TEMPO}^{*}$~\cite{ISPD-2020-TEMPO}} & \multicolumn{2}{c|}{$\text{DOINN}^{*}$~\cite{DAC22-DOINN-Yang}} & \multicolumn{2}{c}{Nitho} \\
                             & MSE           & ME               & PSNR & MSE           & ME               & PSNR & MSE           & ME               & PSNR & mPA         & mIOU       & mPA         & mIOU        & mPA          & mIOU         \\
                             & $\times 10^{-5}$ & $\times 10^{-2}$ & dB   & $\times 10^{-5}$ & $\times 10^{-2}$ & dB   & $\times 10^{-5}$ & $\times 10^{-2}$ & dB   & (\%)        & (\%)       & (\%)        & (\%)        & (\%)         & (\%)         \\ \midrule
      B1                     & 108.29        &10.49           &32.01 &5.55         &1.94 	            &47.10 &\textbf{1.32}          &\textbf{0.51}   &\textbf{50.75} &94.60       &88.70     &99.19	      &98.32	      &\textbf{99.45}	       &\textbf{99.21}         \\
      B2m                    & 1899.04 	     &13.96 	        &30.77 &1202.39 	   &6.11            	&31.64 &\textbf{25.48}         &\textbf{0.82}   &\textbf{49.06} &98.24	      &96.55	   &98.79	      &97.10        &\textbf{99.15}        &\textbf{99.02}         \\
      B2v                    & 6.54          &3.86 	          &42.76 &2.26 	       &2.75 	            &46.37 &\textbf{2.01}          &\textbf{0.68}   &\textbf{48.06} &99.06    	  &93.28  	 &99.21	      &98.41	      &\textbf{99.59}        &\textbf{99.34}         \\
      B2m + B2v                & 4352.25       &15.21           &27.10 &3114.24 	   &12.35             &29.92 &\textbf{33.13}          &\textbf{0.78}   &\textbf{47.88} &98.63    	&95.84  	 &98.71	      &96.68	      &\textbf{99.61}        &\textbf{99.36}         \\ \midrule
      Average                & 1591.53 	     &10.88          	&33.16 &1081.11    	 &5.79            	&39.26 &\textbf{15.49}        	&\textbf{0.70} 	 &\textbf{48.94} &97.63	      &93.59	   &98.98	      &97.63	      &\textbf{99.45}	       &\textbf{99.23}         \\
      Ratio                  & 102.77 	     &15.55         	&0.68  &69.81        &8.27            	&0.80  &\textbf{1.00}          &\textbf{1.00} 	 &\textbf{1.00} &0.98  	    &0.94 	   &0.99 	      &0.98 	      &\textbf{1.00}	       &\textbf{1.00}         \\ \bottomrule
      \multicolumn{16}{l}{\footnotesize{* Models are re-trained using resist image dataset with an amendment to the final activation layer.}} \\
    \end{tabular}}
    \label{tab:results}
\end{table*}

\begin{figure}[tb!]
  \centering
  \includegraphics[width=.8\linewidth]{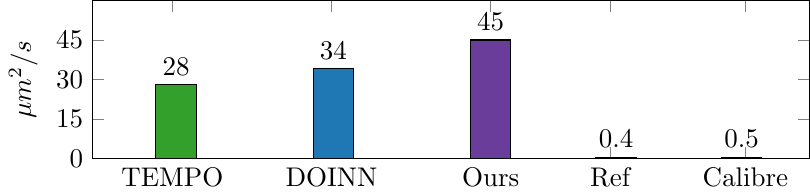}
  \caption{
    Runtime comparison with SOTA.
  }
  \label{fig:runtime}
\end{figure}

We first compare Nitho with TEMPO~\cite{ISPD-2020-TEMPO}, DOINN~\cite{DAC22-DOINN-Yang}, which are SOTA models in aerial and resist stage.
Details are in \Cref{tab:results}, where ``MSE'', ``ME'', ``PSNR'', ``mPA'' and ``mIOU'' are introduced in \Cref{sec:prelim}.
We also merge the B2m and B2v datasets to be ``B2m$+$B2v'' dataset to evaluate our models' performance on larger distribution case.

As can be seen from \Cref{tab:results}, Nitho outperforms SOTA image learning-based models in both aerial and resist stage.
In aerial stage, we achieve 69$\times$ and 102$\times$ smaller MSE than DOINN and TEMPO, with 8$\times$, 15$\times$ smaller ME.
We achieve an average of 48.94dB PSNR compared with 39.26dB of DOINN and 27.10dB of TEMPO.
The results demonstrate the superior advantage of Nitho on high-resolution aerial image generation over previous SOTA.
In resist stage, we achieve above 99\% mPA and mIOU in all datasets with 99.45\%, 99.23\% average mPA and mIOU: 1\% better mPA and 2\% better mIOU than SOTA DOINN.
The improvement can be attributed to the accurate optical kernel regression, which generates high precision aerial images with richer information.
From the results on ``B2m+B2v'' dataset, we find that Nitho's high accuracy can still be maintained, while the performance of DOINN and TEMPO degrades badly in both aerial and resist stage.
It indicates that the image learning-based models have difficulty learning on more complex distributions, while Nitho can still accurately extract optical kernel information.
Result samples are visualized in \Cref{fig:vis_masks}.

\subsubsection{Generalization capability and robustness}
\begin{table}[htbp]
  \centering
  \caption{Comparison with SOTA on out-of-distribution dataset.}
  \setlength{\tabcolsep}{3pt}
	\renewcommand{\arraystretch}{1}
  \begin{tabular}{cc|cc|cc|cc}
  \toprule
  \multicolumn{2}{c|}{Benchmark}                        & \multicolumn{2}{c|}{TEMPO~\cite{ISPD-2020-TEMPO}} & \multicolumn{2}{c|}{DOINN~\cite{DAC22-DOINN-Yang}} & \multicolumn{2}{c}{Nitho} \\
  Train & Test & mPA                     & mIOU        & mPA         & mIOU        & mPA          & mIOU         \\
  on    &  on  & \%                      & \%          & \%          & \%          & \%           & \%           \\ \midrule
  B1   & B1opc & 90.25                   & 86.15       & 98.03       & 94.76       & 99.43        & 99.17             \\
  \multicolumn{2}{c|}{Drop}  & $\downarrow$ 4.35       & $\downarrow$ 2.55  & $\downarrow$ 1.16  & $\downarrow$ 3.56 & $\downarrow$ 0.02 & $\downarrow$ 0.04  \\ \midrule
  B2m  & B2v   & 99.40                   & 71.86       & 99.64       & 78.31            &   99.58           &  97.33            \\
  \multicolumn{2}{c|}{Drop} & $\uparrow$ 0.34       & $\downarrow$ 21.42  & $\uparrow$ 0.43  & $\downarrow$ 20.10 & $\downarrow$ 0.01 & $\downarrow$ 2.01  \\ \midrule
  B2v  & B2m   & 66.06                   & 55.82       & 76.43       & 68.73            &  98.08             &  97.18      \\
  \multicolumn{2}{c|}{Drop} & $\downarrow$ 32.18       & $\downarrow$ 40.73  & $\downarrow$ 22.36  & $\downarrow$ 28.37 & $\downarrow$ 1.07 & $\downarrow$ 1.84  \\ \midrule

  \multicolumn{2}{c|}{Average}   & 85.24  & 71.28  & 91.36   &  80.60           &   \textbf{99.03}   & \textbf{97.90}               \\
  \multicolumn{2}{c|}{Avg. Drop} & $\downarrow$ 12.06  & $\downarrow$ 21.57  & $\downarrow$ 7.70  & $\downarrow$ 17.34 & $\downarrow$ \textbf{0.37} & $\downarrow$ \textbf{1.29} \\
  \bottomrule
  \end{tabular}
  \label{tab:unseen}
\end{table}
In \Cref{tab:unseen} and \Cref{fig:general}(b), we compare three models on out-of-distribution (OOD) datasets to verify the generalization performance.
Column ``train on'' means the model is trained on the referenced dataset but is tested on another dataset with a different mask shape and image distribution from the column ``test on''.
Row ``Drop'' denotes the difference of results between test results on the same distribution and OOD datasets,
with the direction of changes indicated by the up-down arrows $\uparrow, \downarrow$.
The results demonstrate that Nitho can still achieve remarkable accuracy when tested on OOD datasets,
while TEMPO and DOINN suffer strong performance degradation.

\begin{figure}[tb!]
  \centering
  \subfloat[]{\includegraphics[width=.47\linewidth]{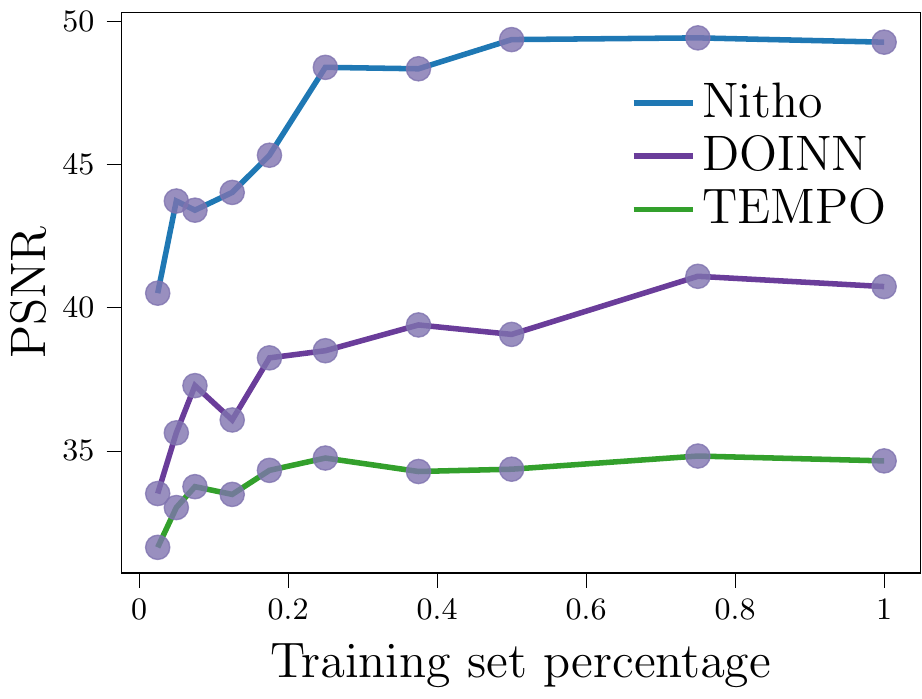} \label{fig:tnum}}
  \subfloat[]{\includegraphics[width=.47\linewidth]{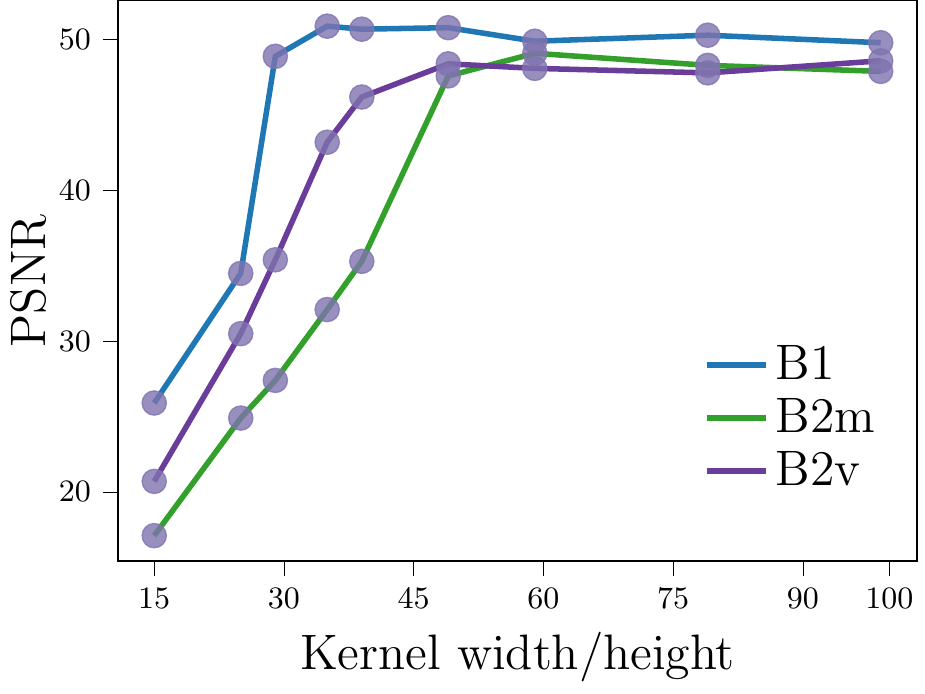} \label{fig:kw}}
  \caption{
    (a) Comparison with SOTA on smaller training sets.
    (b) Ablation study on kernel size on different datasets.
  }
  \label{fig:kw_tnum}
\end{figure}

\Cref{fig:tnum} demonstrates that Nitho can use less training data to achieve better accuracy than previous art.
We list the average PSNR of B1, B2m, and B2v test sets in y-axis of \Cref{fig:tnum}, where the x-axis represents the training set percentage.
It can be concluded that when Nitho only uses 10\% of the training data, it is already more accurate than TEMPO and DOINN with 100\% of the training data.

\subsubsection{Model size and runtime comparison}
As shown in \Cref{tab:diff}, Nitho can use 31\% and 1\% parameters of DOINN and TEMPO to achieve better performance.
\Cref{fig:runtime} shows the runtime comparison of three models and traditional lithography simulators in terms of throughput ($\mu m^2/s$).
With a smaller model size and hierarchical GPU acceleration,
Nitho has 1.3$\times$ and 1.6$\times$ higher throughput than DOINN and TEMPO since no network inference is required.
Compared with traditional lithography simulators,
from which we obtain ground truth aerial and resist images,
Nitho achieves $\sim$90$\times$ speed up with less than 1\% accuracy loss.

\subsection{Ablation Study}
We also conduct experiments to verify our design on kernel dimensions and effectiveness of position encoding.
\subsubsection{Kernel dimensions with resolution limit}
In \Cref{fig:kw}, the x-axis is width/height of the kernel, \ie $(m, n)$ of $\mathcal{K} \in \mathbb{C}^{r \times n \times m}$, and $m = n$ in our settings.
The y-axis is PSNR on the corresponding dataset.
We can observe that as the kernel width/height increase,
the curve flattens out and stops growing.
This indicates that there is an optimal dimension,
and the network can not learn more information after the optimal dimension due to the \textit{resolution limit}.
The experimental results are consistent with the given optimal dimension in \Cref{eq:kernel_mn}.
On the other hand, if $\lambda$ and $N\!A$ of the lithography system are unknown, the optimal dimension can also be obtained through experiments by hyperparameter search.

\subsubsection{Positional encoding}
\begin{table}[htbp]
  \centering
  \caption{Ablation study for positional encoding on B1 dataset.}
  \setlength{\tabcolsep}{2pt}
	\renewcommand{\arraystretch}{1.1}
  \begin{tabular}{l|ccc}
  \toprule
  Type                          & MSE ($\times 10^{-5}$)           & ME ($\times 10^{-2}$) & PSNR (dB)        \\ \midrule
  None                          & 537.32                           & 19.38                 & 25.33            \\
  NeRF PE~\cite{ECCV-2020-NeRF} & 1.79                             & 0.81                  & 48.83            \\
  Ours (\Cref{eq:cplx_gaussian_pe})  & \textbf{1.32}               & \textbf{0.51}         & \textbf{50.75}   \\
  \bottomrule
  \end{tabular}
  \label{tab:pe}
\end{table}
% \subsubsection{Unseen data}
\Cref{tab:pe} illustrates that positional encoding is extremely critical to Nitho.
We first remove the positional encoding layer by using a simple Gaussian matrix.
As shown in the first line of \Cref{tab:pe}, PSNR in test set B1 goes down to 25.33 with worse MSE and ME than TEMPO and DOINN.
Then we apply NeRF's PE in \Cref{eq:nerf_pe} and our RFF PE in \Cref{eq:cplx_gaussian_pe},
we get $\sim$2$\times$ PSNR of 48.83 and 50.75 compared with Nitho without PE.

\section{Conclusion}

In this paper, we abandon the CNN-based image learning methods and devise a new paradigm of lithography model
which separates the influence of mask and lithographic system, aiming to learn optical kernels with neural fields for lithography simulation.
The model starts with deriving the maximum dimension of the optical kernels from physical \textit{`resolution limit'}.
Then Nitho leverages a $\operatorname{\mathbb{C}MLP}$ to learn the implicit mapping from positional encoded coordinates to optical kernel values,
which reflects information from the location-dependent imaging system.
With the predicted kernels, high-precision aerial and resist images  can be obtained following the SOCS formula.
Experiments show Nitho framework outperforms SOTA with a smaller model size in both aerial and resist stages.
At the same time, Nitho demonstrates superior generalization capability than SOTA by a large margin.

%\balance
{
\bibliographystyle{IEEEtran}
\bibliography{ref/Top,ref/DL,ref/DFM,ref/NeRF,ref/bench,ref/CV}
}

\end{document}